\documentclass[manuscript, nonacm]{acmart}
\AtBeginDocument{%
  }

\usepackage[english]{babel}

\usepackage{enumitem}
\usepackage{amsmath}
\usepackage{graphicx}
\usepackage{wrapfig}
\usepackage{hyperref}
\usepackage{caption}
\usepackage{subcaption}
\usepackage{pdflscape}
\usepackage{cleveref}
\usepackage{color, colortbl}
\usepackage{makecell}
\usepackage{multirow}
\usepackage{adjustbox}
\usepackage{tikz}
\usepackage{physics}
\usepackage{amsmath}
\usepackage{tikz}
\usepackage{mathdots}
\usepackage{cancel}
\usepackage{color}
\usepackage{siunitx}
\usepackage{array}
\usepackage{multirow}
\usepackage{gensymb}
\usepackage{tabularx}
\usepackage{extarrows}
\usepackage{booktabs}
\usetikzlibrary{fadings}
\usetikzlibrary{patterns}
\usetikzlibrary{shadows.blur}
\usetikzlibrary{shapes}

\usepackage{pifont}

\newcommand{\conceptexp}{\textit{Post-hoc} concept-based explanation methods}

\newcommand{\conceptmodel}{\textit{Explainable-by-design} concept-based models}
\newcommand{\classconcrel}{{Class-Concept Relation}}

\newcommand{\nodeconcass}{{Node-Concept Association}}
\newcommand{\concvisual}{{Concept Visualization}}

\definecolor{Gray}{gray}{0.9}

\newcommand{\xmark}{\ding{55}}%
\newcommand{\vmark}{\ding{51}}%

\newcommand{\myparagraphbold}[1]{%
  \smallskip
  \textbf{#1.}%
}

\begin{document}

\title{Concept-based Explainable Artificial Intelligence: A Survey}

\author{Eleonora Poeta$^*$}
\email{eleonora.poeta@polito.it}
\author{Gabriele Ciravegna$^*$}
\email{gabriele.ciravegna@polito.it}
\author{Eliana Pastor}
\email{eliana.pastor@polito.it}
\author{Tania Cerquitelli}
\email{tania.cerquitelli@polito.it}
\author{Elena Baralis}
\email{elena.baralis@polito.it}
\affiliation{%
  \institution{Politecnico di Torino}
  \city{Torino}
  \country{Italy}
}


\renewcommand{\shortauthors}{Poeta et al.}

\begin{abstract} 
The field of explainable artificial intelligence emerged in response to the growing need for more transparent and reliable models. However, using raw features to provide explanations has been disputed in several works lately, advocating for more user-understandable explanations.  
To address this issue, a wide range of papers proposing Concept-based eXplainable Artificial Intelligence (C-XAI) methods have arisen in recent years. 
Nevertheless, a unified categorization and precise field definition are still missing.
This paper fills the gap by offering a thorough review of C-XAI approaches. We define and identify different concepts and explanation types. We provide a taxonomy identifying nine categories and propose guidelines for selecting a suitable category based on the development context. Additionally, we report common evaluation strategies including metrics, human evaluations and dataset employed, aiming to assist the development of future methods.
We believe this survey will serve researchers, practitioners, and domain experts in comprehending and advancing this innovative field.    

\end{abstract}



\received{December 2023}

\maketitle

\def\thefootnote{*}\footnotetext{These authors contributed equally to this work}
\def\thefootnote{\arabic{footnote}}

\section{Introduction}

In recent years, the importance of Artificial Intelligence (AI) has surged due to its transformative impact on various aspects of society and industry. This success is predominantly attributed to the advancement of Deep Learning models~\cite{lecun2015deep}. However, the inherent complexity and opaque nature of Deep Neural Networks (DNN) hinders understanding the decision-making process underlying these models. This issue prevents the safe employment of these models in contexts that significantly affect users. 
Consequently, decision-making systems based on Deep learning have faced constraints and limitations from regulatory institutions~\cite{goodman2017european, maccarthy2020examination}, which increasingly demand transparency in AI models~\cite{kaminski2019right}.

To address the challenge of creating more trustworthy and transparent AI models, researchers have pursued eXplainable AI (XAI) methods~\cite{adadi2018peeking}.  Most XAI methods focus on a single prediction, highlighting which input features contributed the most to the prediction of a certain class. This is achieved 
through gradient-based analysis (Vanilla Gradient~\cite{simonyan2014deep}, CAM~\cite{zhou2016learning}, Grad-CAM~\cite{selvaraju2017grad}), or local approximations with surrogate interpretable models (LIME,~\cite{ribeiro2016should}) or game-theoretic approaches (SHAP~\cite{lundberg2017unified}). Other methods, instead, focus on node-oriented explanations to enhance network transparency~\cite{erhan2009visualizing, simonyan2014deep, zeiler2014visualizing}.
Finally, a few methods attempted to globally interpret a model behaviour via global feature importance~\cite{greenwell2018simple} 
linear combination of non-interpretable ones~\cite{agarwal2021neural} or global surrogate models~\cite{ribeiro2016should, guidotti2018local}. 

However, the reliability of standard XAI methods has been contested:~\cite{adebayo2018sanity} showed that gradient-based explanations may not change even after randomly re-parametrizing the network or the data labels, while~\cite{hooker2019benchmark} showed that methods based on surrogate models are unable to identify the most important features.
Furthermore,~\cite{kindermans2019unreliability,ghorbani2019interpretation} 
demonstrated that feature importance methods can be misled by simple data modification, which does not affect the model's prediction. 
Even if these challenges were addressed through the development of more robust methods, a fundamental concern remains: standard XAI techniques provide explanations at the feature level, which may lack meaningful interpretation, especially for non-expert users~\cite{poursabzi2021manipulating}. Understanding \textit{where} the network is looking is not sufficient to explain \textit{what} the network is seeing in a given input \cite{rudin2019stop, achtibat2023attribution}. 
To really achieve this, there is a growing consensus that XAI techniques should provide explanations in terms of higher-level attributes~\cite{kim2018interpretability, achtibat2023attribution, fel2023craft}, commonly referred to as \textit{concepts}.

\begin{figure}[t]
    \centering
    \includegraphics[width=0.6\linewidth]{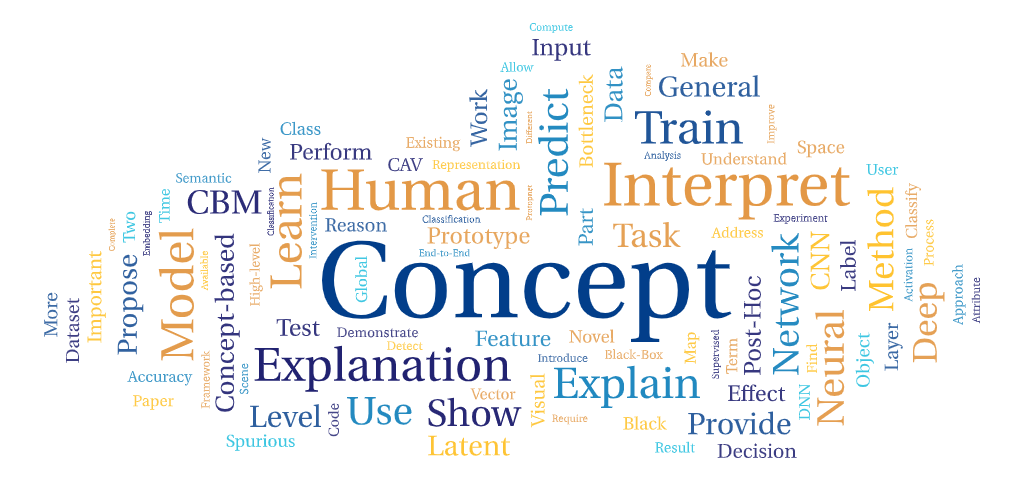}
    \caption{Word cloud generated from the titles, abstracts, and keywords of the reviewed C-XAI papers. The trend towards human understanding and interpretability is noticeable, in contrast to standard XAI approaches.}
    \label{fig:word_cloud}
\end{figure}

Concept-based explanations offer a compelling alternative by providing a more holistic view of the model's inner workings. By explaining the model's predictions in terms of human-understandable attributes or abstractions, concept-based explanations better resemble the way humans reason and explain 
~\cite{kim2018interpretability, kim2023help}. 
 They also empower users to gain deeper insights into the underlying reasoning, helping them to detect model biases and improve the classification model~\cite{zhou2018interpretable, jain2022extending, bontempelli2023conceptlevel}. Finally, they are more stable to perturbations than standard XAI techniques~\cite{alvarez2018towards} and can create models inherently more robust to adversarial attacks~\cite{ciravegna2023logic}.

This paper proposes the first systematic review of Concept-based eXplainable Artificial Intelligence (C-XAI) methods. 
While there exists an abundant literature of surveys in the realm of explainability~\cite{guidotti2018survey, adadi2018peeking, das2020opportunities, arrieta2020explainable, dwivedi2023explainable, bodria2023benchmarking, poche2023natural}, none of them has distinctly concentrated on the concept-based domain and only few~\cite{bodria2023benchmarking, poche2023natural} identified C-XAI as a separate category.
We analyzed papers from 2017 to July 2023 appearing in the proceedings of NeurIPS, ICLR, ICML, AAAI, IJCAI, ICCV and CVPR conferences or appeared in important journals of the field such as Nature Machine Intelligence, Transaction on Machine Learning Research, Artificial Intelligence, or also cited in the related work of these papers. 
To gain a visual insight into the examined papers, please refer to Figure~\ref{fig:word_cloud}. This illustration showcases the 100 most frequently occurring words found in the titles, abstracts, or keywords of the papers under review. The figure already indicates a noteworthy trend in the C-XAI field, where the emphasis is shifting towards interpretable models and explanations more in line with human understanding.

To summarize, this work contributes with the following:
\begin{itemize}
    \item We properly define the key terms in the concept-based explanation field, including what is a concept and a concept-based explanation.
    \item We present a taxonomy that outlines 9 different C-XAI categories based on the utilization of concepts during training, the specific type of concept used, and the required explanation.
    \item We propose guidelines for choosing a suited approach, describing each category's applicability, advantages, and disadvantages.
    \item We analyze the methods according to 13 dimensions of analysis.
    \item We provide for each method a brief outline of its main aspects and how it differs from the other methods in the same category, including papers of critics and comparisons.
    \item We report metrics and datasets available for assessing and developing novel C-XAI methods.
    \item We highlight first C-XAI applications and promising future directions.
\end{itemize}

We hope this paper may serve as a reference for AI researchers and practitioners seeking to enhance the reliability of explanation methods and models.
The suggested guidelines are a valuable tool for domain experts, enabling them to select the most fitting explanation method tailored to the specific type of explanation needed or the concepts at hand, for instance.
Policymakers may also find concept-based models to be sufficiently accountable and trustworthy for the deployment of neural models in safety-critical contexts, 
making this paper relevant to a wide spectrum of stakeholders.    

The paper is structured as follows: in Section~\ref{sec:def_and_cat} we provide the definitions of the most important terms in the concept-based explanation field; in Section~\ref{sec:categorization}, we define a high-level categorization of C-XAI approaches and provide readers with guidelines for selecting the suitable category; in Section~\ref{sec:concept_exp} we resume the most important post-hoc concept-based explainability methods, while in Section~\ref{sec:concept_model} we review explainable-by-design concept-based models following the previously introduced categorization; 
in Section~\ref{sec:evaluation}, we report valuable tools developed in the literature to develop and assess novel methods, including metrics, human evaluations, and datasets; 
in Section~\ref{sec:application}, we report the first applications of concept-based explainability methods;
in Section~\ref{sec:future_work}, we hypothesize some promising future directions of the concept-XAI field; 
finally, Section~\ref{sec:conclusions} concludes the paper. 

\section{Definitions} 
\label{sec:def_and_cat}
\begin{figure}
    \centering
    \includegraphics[scale=0.33]{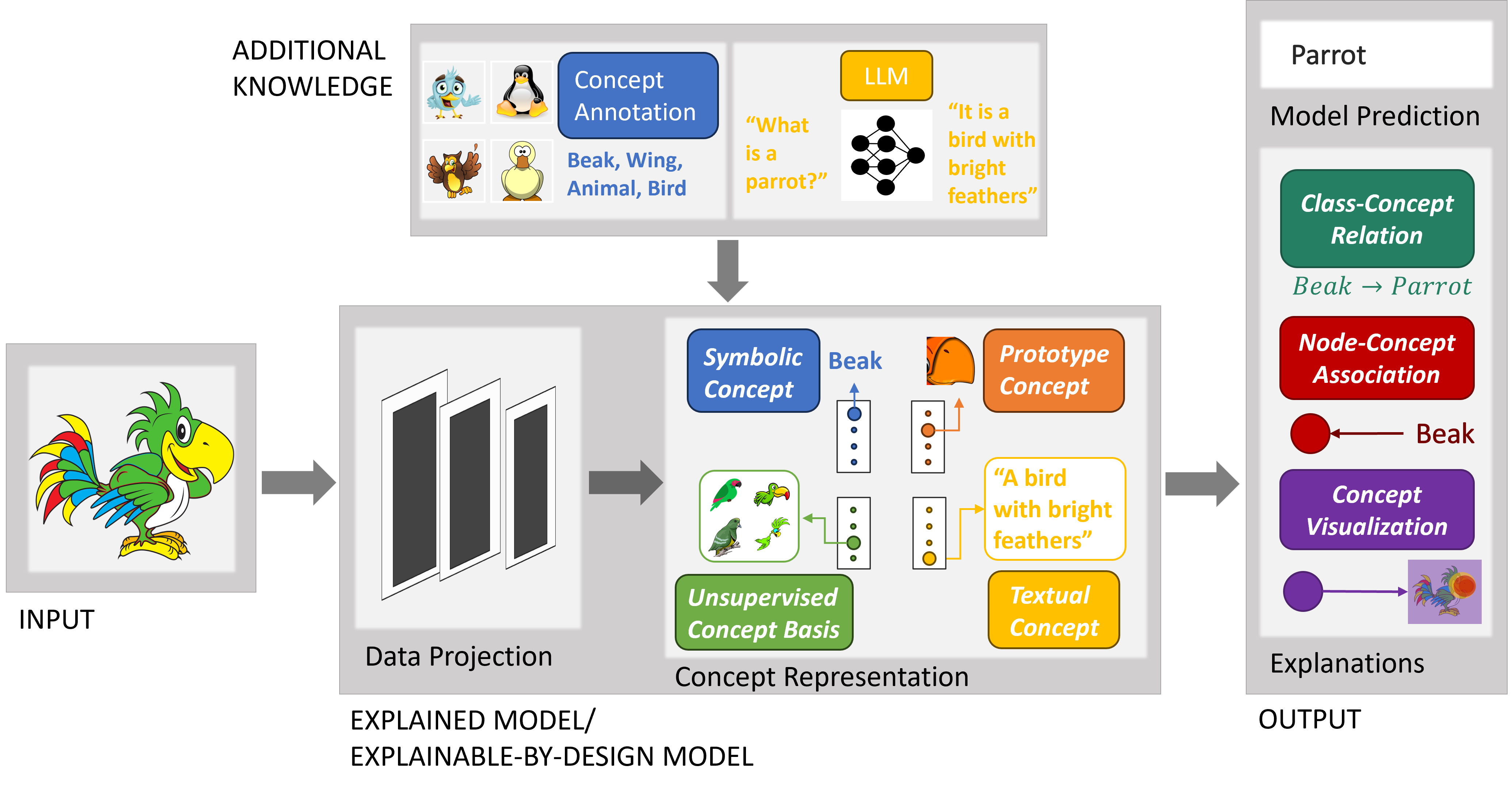}
    \caption{Concepts and explanations provided by C-XAI methods and models. Some concepts require additional knowledge to be implemented.}
    \label{fig:vis_abstract}
\end{figure}

In Figure~\ref{fig:vis_abstract}, we provide a common pipeline of a C-XAI method. The input is projected in the latent space of the network, where the method aims to extract or directly represent a set of concepts (normally of the same type). For some concepts, additional knowledge may be required to employ them properly. Finally, various types of explanations can be generated in addition to the model prediction. 
Due to the recent genesis of the C-XAI literature, it is challenging to establish consensus on the definitions of certain terms. Specifically, the definitions of “concept” and “explanation” are not consistently uniform, and similar studies may adopt varying interpretations.

In the following, we first propose four typologies of concepts (Section~\ref{sec:what_concept}); secondly, we identified three different concept-based explanations with different types of explainability/interpretability (Section~\ref{sec:expltype}). Further, we recall the difference between a post-hoc method and an explainable-by-design model, particularly within the C-XAI context (Section~\ref{sec:post_vs_cbm}). Finally, in Appendix~\ref{app:glossary}, we compiled a glossary of the most commonly used terms, while in Table~\ref{tab:acronyms}, we report the acronyms and abbreviations used throughout the paper.

\subsection{What is a Concept?}
\label{sec:what_concept}
In the concept-based explainability literature, the term \textit{concept} has been defined in very different ways. Citing~\cite{molnar2020interpretable}, “A concept can be any abstraction, such as a colour, an object, or even an idea”. 
As we resume in the top part of Table~\ref{tab:conc_exp_types}, we propose a categorization of concepts into four typologies: 
\textit{Symbolic Concepts}, \textit{Unsupervised Concept Bases}, \textit{Prototypes}, and \textit{Textual Concepts}. 

\textbf{\textit{Symbolic Concepts}} are human-defined symbols~\cite{ciravegna2023logic}. They can be high-level attributes of the task under consideration or interpretable abstractions, such as the color or the shape of the predicted object~\cite{barbiero2022entropy}. Figure \ref{fig:vis_abstract} shows that \textit{beak} is a suitable symbolic concept for a bird identification task. Since these concepts are pre-defined by humans, we generally require auxiliary data equipped with concept annotation, particularly when we are dealing with non-symbolic features (e.g., image pixels vs tabular data). The representation in the network of symbolic concepts can be analyzed post-hoc or forced during training to create an explainable-by-design model.

\textbf{\textit{Unsupervised Concept Bases}} are clusters of samples the network learns. Even if they are not built to resemble human-defined concepts, these unsupervised representations may still capture abstractions that are more understandable to humans than individual features or pixels~\cite{ghorbani2019towards}. As shown in Figure~\ref{fig:vis_abstract}, a network may learn a cluster of green birds. To extract such a concept, we require the employment of a clustering algorithm either post-hoc in the latent space of the model or during training, e.g., by means of a reconstruction loss. As in the previous case, these concepts can be employed by both explainable-by-design models or post-hoc methods.

\textbf{\textit{Prototypes}} are representative examples of peculiar traits of the training samples. They can be a training sample or only a part of a training sample. As shown in Figure~\ref{fig:vis_abstract}, a part prototype of a training sample might be a \textit{beak}. The set of prototypes is, in general, representative of the whole data set~\cite{li2018deep}. We regard prototypes as concepts following recent literature~\cite{marconato2022glancenets} since they are still higher-level terms than the input features. Also, they are different from unsupervised concept bases since they explicitly encode the peculiar traits in the network weights, and consequently, they cannot be employed in post-hoc methods but only in explainable-by-design models. 

\textbf{\textit{Textual Concepts}} are textual descriptions of the main classes. As reported in Figure~\ref{fig:vis_abstract}, a textual concept might be `A bird with bright feathers' for a sample of a parrot. Textual concepts are provided at training time by means of an external generative model~\cite{oikarinen2023label}, and they are employed inside a concept-based model in the form of a numerical embedding of the text. This type of concept is gaining prominence due to the recent development of Large-Language Models (LLMs). So far, it has been only employed in explainable-by-design models.

For a more formal definition of a Concept, following category theory principles, please refer to Appendix~\ref{sec:formal_concept}.

{%
\renewcommand{\arraystretch}{1.05}
\begin{table}[t]
    \centering
    \begin{tabular}{l|l}
        \hline
        \rowcolor{Gray}
        Concept Type                & Definition \\
        Symbolic Concept            & Human-defined attribute or abstraction \\
        Unsupervised Concept Basis  & Cluster of similar samples\\
        Prototype                  & (Part-of) a training sample\\
        Textual Concept            & Short textual description of a main class \\
        \hline
        \rowcolor{Gray}
        Explanation Type                &  \\
        Class-Concept Relation    & \begin{tabular}[c]{@{}l@{}}Relation among a concept and \\ an output class of a model \end{tabular}     \\
        Node-Concept Association & \begin{tabular}[c]{@{}l@{}}Explicit association of a concept with \\ a hidden node of the network\end{tabular} \\
        Concept-Visualization  & \begin{tabular}[c]{@{}l@{}}Visualization of a learnt concept in terms \\ of the input features\end{tabular}   

    \end{tabular}
    \caption{Concept and explanation types.}
    \label{tab:conc_exp_types}
\end{table}
}

\subsection{What is a Concept-based Explanation?}
\label{sec:expltype}
Even when considering a fixed type of concept, the definition of a concept-based explanation can be elusive. Existing literature generally agrees that concept-based explanations should explain how DNNs make particular decisions using concepts~\cite{yeh2020completeness} and adhere to specific criteria such as being meaningful, coherent, and relevant to the final class~\cite{ghorbani2019towards} and also explicit and faithful~\cite{alvarez2018towards}. 
As we report in the bottom part of Table~\ref{tab:conc_exp_types}, we define three distinct categories of concept-based explanations: \textit{\classconcrel}, \textit{\nodeconcass}, and \textit{\concvisual}. Each method may offer one or more explanations, serving specific purposes in different scenarios.

\textbf{\textit{\classconcrel}} explanations consider the relationship between a specific concept and an output class within the model. This relationship can be expressed either as the importance of a concept for a particular class or through the use of a logic rule that involves multiple concepts and their connection to an output class. For example, in Figure~\ref{fig:vis_abstract}, the image in input is classified as a Parrot because it is mainly related to the concept of \textit{beak}. In general, this type of explanation can be applied to all types of concepts: for instance, with prototypes, we have class-concept explanations like $parrot :=  0.8 \  prototype_1 + 0.2 \ prototype_2$. This explanation can be extracted by analyzing post-hoc the correlation of concepts and classes in the latent space of a network or with the employment of an interpretable model from the concepts to the tasks.  

\textbf{\textit{\nodeconcass}} explicitly assigns a concept to an internal unit (or a filter) of a neural network. This explanation enhances the transparency of deep learning models, highlighting what internal units see in a given sample. 
It can be defined post-hoc by considering the hidden units maximally activating on input samples representing a concept. 
Otherwise, it can also be forced during training by requiring a unit to predict a concept.  
For instance, consider Figure\ref{fig:vis_abstract}: some nodes of the network can automatically learn to recognize the beak since it is a crucial part of the prediction of a bird. Otherwise, an expert can also define a set of attributes in advance and require their explicit representation in the network intermediate layers.

Finally, \textbf{\textit{\concvisual{}}} explanations highlight the input features that best represent a specific concept. When symbolic concepts are used, this explanation closely resembles the saliency map of standard XAI methods. However, when non-symbolic concepts are employed (such as unsupervised concept basis, prototypes, or even textual concepts), the focus shifts towards understanding which unsupervised attributes or prototypes the network has learned. This form of explanation is often combined with one of the previous explanations, and it enables the understanding of what are the concepts associated with a specific class or node. 
For instance, consider again Figure~\ref{fig:vis_abstract}, where we reported a model employing prototype concepts as an example of \concvisual{}. A visualization of the prototype in the input sample 
is crucial to understand which concept the network has learned. 

\subsection{C-XAI Post-hoc Methods vs Explainable-by-design Models}
\label{sec:post_vs_cbm}
In the XAI literature, the difference between \textit{post-hoc explanation methods} and \textit{explainable-by-design models} is well-defined: 
the latter are inherently designed for transparency, such as decision trees or linear models; in contrast, the former are tools applied post-training to explain complex models, such as DNNs. 
On the contrary, in the field of concept-based approaches, the difference becomes more subtle since C-XAI explainable-by-design models also work with DNNs but explicitly represent some concepts within the architecture. In the following, we try to clarify this difference.  

\textbf{\textit{Post-hoc Concept-based Explanation Methods}} operate on concepts rather than dissecting input features. They explain a prediction, an entire class, or an internal network node given a set of concepts. Typically, this involves projecting the samples representing the concepts in the model's latent space and analyzing their relation to the prediction and the hidden node activations. 
Symbolic and unsupervised concept bases have been both employed to provide all types of explanations.

\textbf{\textit{Explainable-by-design Concept-based Model}}, on the other hand,
are neural models employing an explicit representation of a set of concepts as an intermediate layer. In this way, the predicted concepts influence the task predictions, and thus, they are closely tied to the provided explanations. Since they generally provide Node-Concept association by-design, they can be regarded as inherently transparent models, at least to some extent.
In this case, the whole spectrum of concepts can be used, either explicitly annotated, extracted in an unsupervised manner, encoding prototypes, or even generated by a separate model.

\section{C-XAI Taxonomy} 
\label{sec:categorization}
To assist users in understanding the C-XAI landscape, in this section, we provide a taxonomy of concept-based methods and models together with some guidelines for selecting an appropriate method for the development constraints and the desired outcomes. 
In Section~\ref{sec:dimanalysis}, we describe the dimensions of analysis that we employed to categorize and characterize C-XAI methods,
while in Section~\ref{sec:guidelines}, we present a procedure to follow to select the desired C-XAI method for different requirements. 

\subsection{Dimensions of analysis}
\label{sec:dimanalysis}
We characterize each concept-based explainability method by 13 dimensions of analysis. 
We categorize these dimensions into three groups: (i) concepts and explanations characteristics, (ii) the applicability of the method, and (iii) resources employed. The dimensions related to the first two groups will be analyzed in Sections~\ref{sec:concept_exp},~\ref{sec:concept_model} and reported in Tables~\ref{tab:summary:posthoc1},~\ref{tab:summary:bydesign1}, respectively analyzing methods following two different concept training (first dimension of analysis). The third group of dimensions will be analyzed vertically (per resource/evaluation) in Section~\ref{sec:evaluation}, but we also report a horizontal analysis (per method) in Section~\ref{sec:resources}, and in the Appendix in Tables~\ref{tab:summary:posthoc2} and~\ref{tab:summary:bydesign2}.  




\begin{enumerate}
    \item \textbf{Concept and Explanations Characteristics}.  These dimensions capture key aspects of concepts and explanations, such as how concepts are integrated into models, concept annotation strategies, concept types, the form of explanations, and their scope.
        \begin{itemize}
            \item the \textit{Concept Training}, an approach is first categorized according to whether it employs concepts only to provide explanations of an existing model (Post-hoc methods), or also while training the same model (Explainable-by-design models);
            \item the \textit{Concept Annotation}, according to whether the method employs an annotated set of concepts (Supervised), it does not (Unsup.), it employs only a few concepts (Hybrid), or it generates concept annotation (Generative);
            \item the \textit{Concept Type} (described in Section~\ref{sec:what_concept}), which can be either symbolic (Symb.),  unsupervised concept bases (Uns. Basis), prototype-based (Proto.), or textual;
            \item the \textit{Explanation Type}, (defined in Section~\ref{sec:expltype}), either Class-concept relation (C-CR), Node-concept association (N-CA), Concept Visualization (C-Viz) or a combination of these;
            \item the \textit{Explanation Scope}, according to whether the explanation is local (LO) (i.e., it explains an individual prediction), global (GL) (i.e., it provides insights into the overall model behavior), or both.
        \end{itemize}
        Regarding Concept-based Models, 
        we also characterize the methods with the following:
        \begin{itemize}
            \item the \textit{Concept Employment}, how an explainable by-design supervised model employs the concepts during training, either through joint training (Joint) or concept instillation (Instil);
            \item the \textit{Performance Loss}, whether 
            there is a performance loss (\vmark) or not (\xmark) compared to a black-box model.
        \end{itemize}

    \item \textbf{Applicability of the approach}. These dimensions describe the approach's applicability by detailing the types of data it can handle, the primary tasks it is designed for, and the specific neural architectures utilized.
        \begin{itemize}
            \item the \textit{Data}, i.e., the data type the method can handle, generally images (IMG), but also text (TXT), graph (GRA), tabular data (TAB), videos (VID) and time series (TS);
            \item the \textit{Task} the method is designed for, primarily classification (CLF) but also regression (REG) and clustering (CLU);
            \item the \textit{Network} dimension describes the type of neural network employed, including Convolutional Neural Network (CNN),  3D-CNN, Auto-Encoder (AE), Graph Neural Network (GNN), Multi-Layer Perceptron (MLP) and Transformer (TRANSF).
        \end{itemize}
    \item \textbf{Resources and Evaluation conducted}. These dimensions focus on the resources provided (data releases and code/models availability) and the conducted evaluation.
        \begin{itemize}
            \item the \textit{Data Release} indicates whether a new dataset annotated with concepts has been made available;
            \item The \textit{Code/Models Availability} indicates whether the authors have released the code and/or the model;
            \item the \textit{New metric} dimension indicates if the paper introduces a new evaluation metric;
            \item the \textit{Human evaluation} dimension indicates whether the paper includes a user study to evaluate the method.
        \end{itemize}   
\end{enumerate}

\begin{figure}[t]
    \centering
    \resizebox{0.85\textwidth}{!}{%
    \input{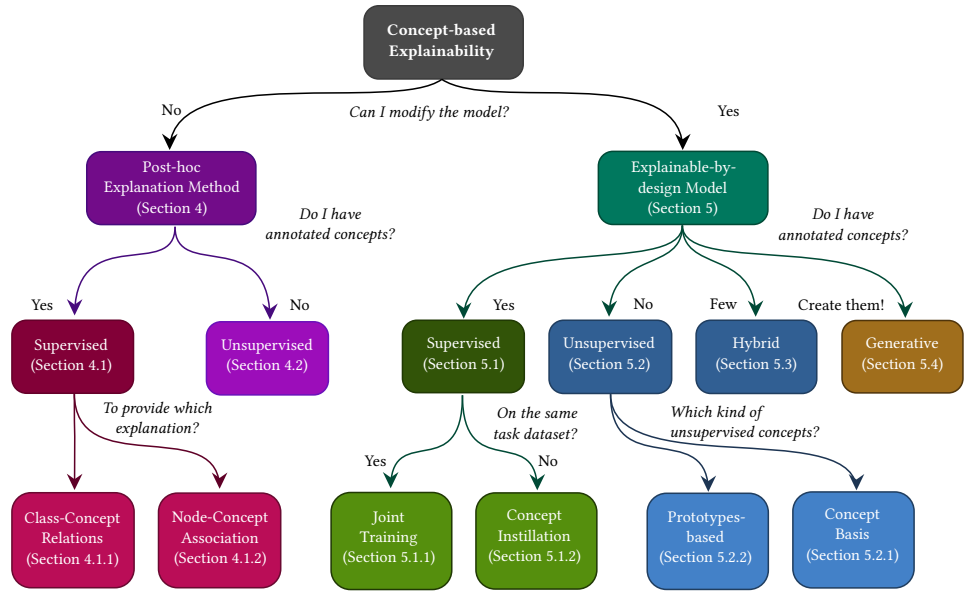}
    }
    \caption{Categorization of the C-XAI methods with guidelines for selecting a suitable approach. 
    }
    \label{fig:categorization}
\end{figure}

\subsection{Guidelines for selecting a suitable method}
\label{sec:guidelines}
We now provide some guidelines to facilitate the method selection, employing Figure~\ref{fig:categorization} as a visual reference.
In particular, let us consider a user interested in a C-XAI method. We propose a set of a maximum of three questions that he should answer to choose a category of methods. These questions regard the possibility of modifying the learning model, the availability of an annotated set of concepts, and, according to the subcategory, the desired outcome regarding explanation or concept types or the type of training set employed. We adopted a conversational approach to help the user identify a solution that is suitable for his development conditions. 
As we report in the figure, this categorization will also characterize the structure of the following sections.

\textbf{\textit{Can I modify the model?}} 
This represents the first coarse-grained question a user should answer. According to the possibility of intervening in the model's development, he can choose among two macro-categories in which the C-XAI literature is divided.
He must look at \conceptexp{} (Section~\ref{sec:concept_exp}) if he needs to consider an already trained model that he cannot (or does not want to) modify. On the other hand, if he can create a model from scratch, he can employ a \conceptmodel{} (Section~\ref{sec:concept_model}), which allows an explicit representation of the concepts within the model architecture.

\textbf{\textit{Do I have annotated concepts?}} Secondly, the user should check whether he can find a dataset annotated with concepts that are related to the task at hand.  
In case of a positive response, he should look at supervised methods; otherwise, he should consider unsupervised ones that can extract concepts from the same data automatically.
For explainable-by-design approaches, he can pick from two more categories. He can employ a hybrid solution that leverages a few supervised concepts together with unsupervised extracted ones since, recently, he can also use a generative method that creates the concepts by means of an external model. 

For some sub-categories in Figure~\ref{fig:categorization}, we identify a third-level of fine-grained categories (and associated questions) according to whether the explanation, the concept employed, or the concept annotation varies among the methods within the subcategory. 
\textbf{\textit{To provide which Explanation?}} - If he selected a post-hoc supervised approach, these differ in the type of explanations, either class-concept relations or node-concept associations.
\textbf{\textit{On the same task Dataset?}} - Explainable-by-design supervised approaches either require concept annotation on the same dataset of the task at hand or allow them on a separate one. 
\textbf{\textit{Which kind of unsupervised concepts?}} - Finally, if he chose an unsupervised explainable-by-design approach, they differ in the representation of unsupervised concepts adopted. Instead, the methods in the remaining sub-categories share these main characteristics; hence, we have not split them further.

\section{\conceptexp{}} 
\label{sec:concept_exp}
\begin{table}[t]
\small
\caption{Post-hoc Concept-based Explainability methods. 
We characterize the approaches based on the dimensions describing concepts and their explainability and the applicability of the approach.
    A full description of each category and of the acronyms is provided in Section~\ref{sec:dimanalysis}.
    }
    \label{tab:summary:posthoc1}
\begin{tabular}{ll|l|cccccc}
 & & \multicolumn{1}{c|}{\textbf{Method}} & \textbf{\begin{tabular}[c]{@{}c@{}}Concept \\ Type\end{tabular}} & \textbf{\begin{tabular}[c]{@{}c@{}}Expl.\\ Type\end{tabular}} & \textbf{\begin{tabular}[c]{@{}c@{}}Expl.\\Scope\end{tabular}}  & \textbf{\begin{tabular}[c]{@{}c@{}}Data \\ Type\end{tabular}} & \textbf{Task} & \textbf{\begin{tabular}[c]{@{}c@{}}Network \\ Type\end{tabular}} \\ 
\hline
\multirow{14}{*}{\rotatebox[origin=c]{90}{\textbf{Concept Annotation}\hspace{1cm}}}
& \multirow{9}{*}{\rotatebox{90}{Supervised}} 
& T-CAV   \cite{kim2018interpretability}        & \multirow{5}{*}{\rotatebox{0}{Symb.}} & C-CR         & GL        & IMG & CLF & CNN \\
& & CAR \cite{crabbe2022concept}                &  & C-CR         & GL        & IMG + TS & CLF & CNN \\
& & IBD \cite{zhou2018interpretable}  &  & C-CR, C-Viz  & LO \& GL  & IMG & CLF & CNN \\ 
& & CaCE \cite{goyal2019explaining}             &  & C-CR         & GL        & IMG & CLF & CNN \\ 
& & CPM \cite{wu2023causal}                     &  & C-CR         & LO        & TXT & CLF & TRANSF \\
\cline{3-9}
& & Obj. Det.  \cite{zhou2015object}            & \multirow{4}{*}{\rotatebox{0}{Symb.}} & N-CA, C-Viz       & GL & IMG & CLF & CNN \\
& & ND  \cite{bau2017network}                   &  & N-CA, C-Viz       & GL & IMG & CLF & CNN \\
& & Net2Vec \cite{fong2018net2vec}              &  & N-CA, C-Viz       & GL & IMG & CLF & CNN \\
& & GNN-CI                                      &  & N-CA, C-CR, C-Viz & GL & GRAPH & CLF & GNN \\
 \cline{2-9}
& \multirow{7}{*}{\rotatebox[origin=c]{90}{Unsupervised}} 
& ACE \cite{ghorbani2019towards}                & \multirow{6}{*}{\rotatebox{0}{C.Basis}} & C-CR, C-Viz   & GL         & IMG & CLF & CNN \\
& & Compl. Aware \cite{yeh2020completeness}     &   & C-CR          & GL         & IMG, TXT    & CLF & CNN    \\
& & ICE \cite{zhang2021invertible}              &   & C-CR, C-Viz   & LO \& GL   & IMG & CLF & CNN \\
& & MCD \cite{vielhaben2023multi}               &   & C-CR, C-Viz   & LO \& GL   & IMG & CLF & CNN, TRANSF \\
& & CRAFT   \cite{fel2023craft}                 &   & C-CR          & GL         & IMG & CLF & CNN \\ 
& & DMA \& IMA \cite{leemann2023are}            &    &   -           &   -        &  IMG   &   CLF & CNN  \\
& & STCE   \cite{ji2023spatial}                 &   & C-CR, C-Viz   & GL         & VID  & CLF & 3D-CNN  \\
\hline
\end{tabular}
\end{table}

Concept-based explanation methods explain an existing model without modifying its internal architecture.  For this reason, they are post-hoc methods, i.e., they analyze the model only after it has finished the training process. 

As we can observe from Table~\ref{tab:summary:posthoc1}, 
\conceptexp{} primarily diverge on the employment of a dataset containing concept annotations (Supervised, Section~\ref{sec:sup_concept_exp}) or not (Unsupervised methods, Section~\ref{sec:unsup_concept_exp}). 
While in the first case, they employ Symbolic Concepts, in the latter, they extract explanations based on unsupervised concept basis. 
Furthermore, supervised methods differ in focusing on either explaining class-concept relations (C-CR) or node-concept associations (N-CA). In both cases, they can also provide concept visualizations (C-Viz). Unsupervised methods, instead, generally focus on class-concept relations as well as on visualizing the extracted concepts. The explanation scope of most methods is global (GL), with few methods providing both local and global explanations (LO\&GL) and one only local (LO). 
All the reviewed methods have been developed to explain classification tasks (CLF), and many have been designed for CNNs and image (IMG) data types.  
A few methods have been developed for working with texts (TXT), graphs (GRAPH), or videos (VID) and employing transformers (TRANS) (both for texts and images), GNNs, and 3D-CNN architectures. 
To characterize the key aspects of post-hoc methods, here we discuss the main positive aspects versus the negative ones.

\myparagraphbold{Advantages}
Post-hoc explanation methods are the only viable option in scenarios where a pre-trained model exists or a predefined model is necessary.
Also, Concept-based explainability methods are the preferred choice when there can be no compromise on the predictive and generalization capabilities of the model. They allow for enhanced interpretability without affecting the model's performance.  

\myparagraphbold{Disadvantages}
The main drawback of these methods is that they do not guarantee that the model truly comprehends or employs the adopted concepts. This is because the model was not exposed to these concepts during its training process. Concept-based explainability methods can only be regarded as a better way of explaining the network behavior by employing terms that are more comprehensible to humans than raw input features. We will further discuss the issues and challenges post-hoc methods face at the end of this section (Section~\ref{sec:posthoc_issues}).


\subsection{Supervised Concept-based Explainability Methods} \label{sec:sup_post_hoc}
Post-hoc supervised concept-based explanation methods analyze network behavior over samples annotated with \textit{symbolic concepts}. The underlying idea is that a network automatically learns to recognize some concepts. These methods aim to assess which concepts have been learned, where, and how they influence the model. 
This analysis is conducted differently according to the type of explanations delivered. 
In particular, to provide Class-concepts relations, some methods analyze how the current prediction or an entire class weight correlates with the projection of a concept in the latent space of the model (Section \ref{sec:post-sup-class}). On the contrary, methods extracting explanations in terms of Node-concept associations study the activations of single hidden nodes to associate them to a given concept (Section \ref{sec:post-sup-node-concept-assoc}). 

\subsubsection{\classconcrel{} methods}
\label{sec:post-sup-class}
\begin{figure}[t]
    \centering
    \includegraphics[scale=0.33]{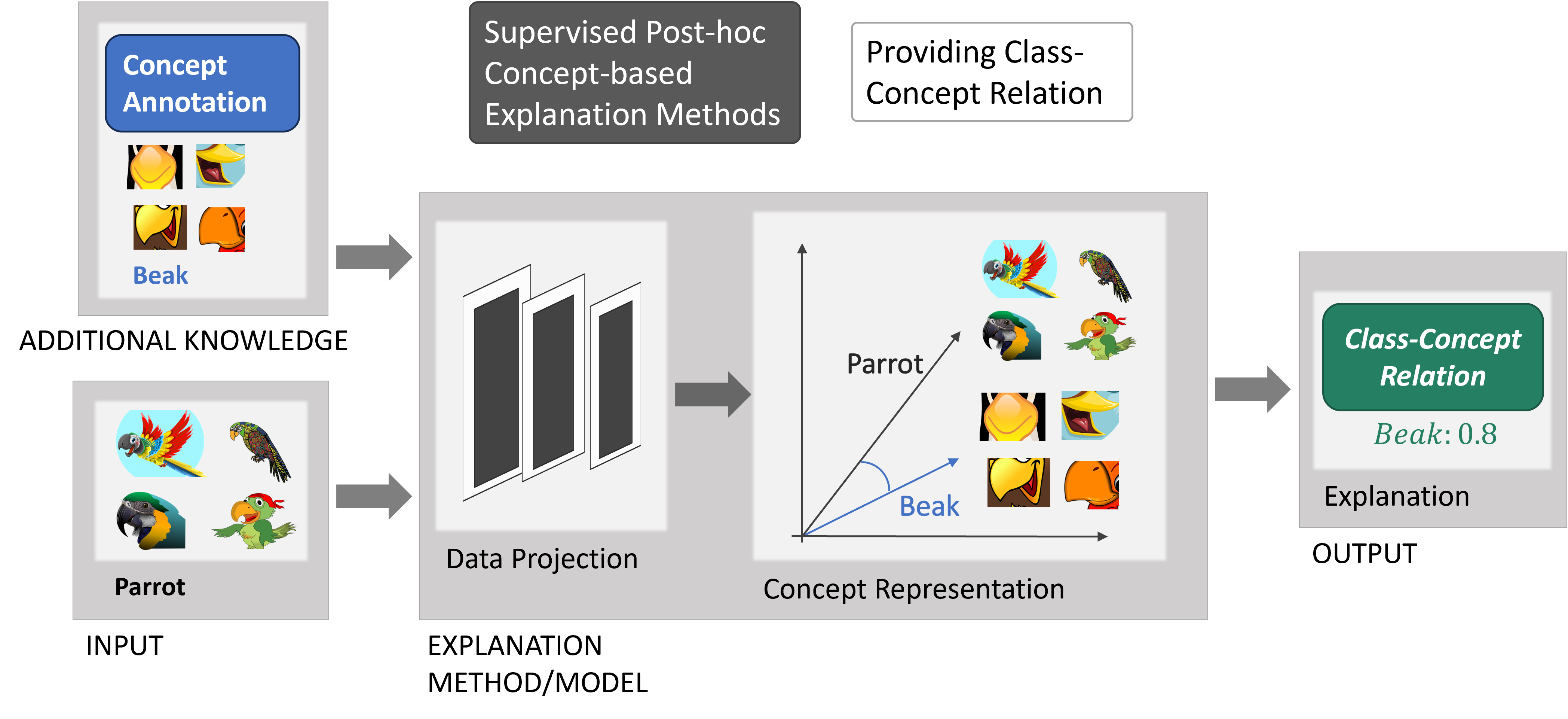}
    \caption{Post-hoc Supervised Methods providing explanations in terms of class-concept relations. 
    These approaches provide a set of samples annotated with concepts to the explained network to determine their influence on the output class.
    }
    \label{fig:post_hoc_sup_class_conc}
\end{figure}

Supervised post-hoc methods offering explanations examine the connection between output classes and a set of symbolic concepts. As illustrated in Figure~\ref{fig:post_hoc_sup_class_conc}, 
this is generally achieved by presenting samples representing concepts to the explained network and analyzing its behaviour. In the figure, we report the case of a model trained to recognize birds (e.g., a Parrot) and tested with samples representing a related concept (e.g., the Beak). The effect of the concepts on the final class is studied by either (i) analyzing the representation of concepts in the model's latent space and its relation with the output classes (TCAV, TCAR, IBD) or (ii) exploring the causal effect of inserting or removing a concept on the prediction of a particular class (CaCE, CPM). 
The samples representing the symbolic concepts are drawn from an external dataset, which is then a requirement for this class of methods. As previously mentioned, the set of concepts must be related to the output classes to extract meaningful explanations.

\smallskip
Testing with Concept Activation Vector \textbf{(T-CAV~\cite{kim2018interpretability})} has been the first work in this category.
It introduces the idea of employing a linear probe to assess the interpretability of a model. It consists of (i) freezing the model's internal architecture and (ii) training a linear layer in the latent space of the model to predict the concepts.
For each user-defined concept, it requires 
a positive set comprising images representing the concept and a complementary one with negative samples. 
The Concept Activation Vectors (CAVs) represent the orthogonal vectors to the weights of the linear probe, separating positive examples of a given concept from negative ones. 
Then, T-CAV computes the inner product between the directional derivatives of the class and the CAVs to quantify the models' conceptual sensitivity, i.e., 
how much each concept positively influences the prediction of a given class. 
The T-CAV score is the fraction of the class's inputs with a positive conceptual sensitivity. The higher the score, the higher the influence of the concept.
As noted by the authors, CAVs require concepts to be linearly separable. However, concepts may be semantically linked and mapped to latent space portions that are close to each other. 
In this case, the authors propose to employ Relative CAVs, i.e., CAVs obtained when considering as negative samples the positive concepts of other concepts. 
The authors showed that T-CAV enables humans to identify the most important concepts for a network trained over standard and corrupted features, while several standard XAI methods do not.

\smallskip
Concept Activation Regions \textbf{(CAR~\cite{crabbe2022concept})} relaxes the linear separability assumption of concepts imposed by T-CAV. 
CAR only requires concept examples to be grouped in clusters in the model's latent space such that they are identifiable by a non-linear probe (e.g., a kernel-based SVM). Each concept is thus represented by a \emph{region} in the latent space and no longer by a single vector. 
To this aim, the authors define the notion of concept smoothness, i.e., the possibility to separate a \textit{positive region} in which the concept is predominantly present and a \textit{negative region} in which the concept is not present.
A concept is important for a given example if its projection lies in the cluster of concept positive representations.
Equivalently to the TCAV score, the authors propose the TCAR score, quantifying the proportion of instances belonging to a specific class whose representations fall within the positive concept region. 
In some settings, TCAR returns high scores expressing significant associations between classes and concepts, while TCAV returns low scores. 
The authors claim that this is due to employing a more powerful concept classifier and, consequently, that TCAR scores reflect the relations between classes and concepts more accurately.

\smallskip
Interpretable Basis Decomposition \textbf{(IBD~\cite{zhou2018interpretable})} 
provides class-concept relations by decomposing the evidence supporting a prediction into the most semantically related components. 
Concretely, IBD considers the classes and the concepts as vectors in the latent space of the penultimate layer of the network.
As for T-CAV, to represent the concepts, IBD trains a set of linear classifiers over the projections in the latent space. 
IBD then performs an optimization to assess the relevance of each concept for a class. For all samples in the concept dataset, they try to reconstruct a given class's prediction using a linear regression. In this, the variables are the inner product between each concept vector and the projection of the sample in the latent space. Instead, the coefficients represent each concept's associated importance for the given class.
Interestingly, the \emph{residual} of the optimization assesses the variability in the class predictions that cannot be represented by means of the learned concepts.
Furthermore, each concept is accompanied by a heat map showing the features with the highest impact on the concept prediction. 
By means of human evaluation, the authors demonstrated that IBD explanations enhance standard XAI techniques (like CAM and Grad-CAM) regarding model assessment quality.

\smallskip
The Causal Concept Effect \textbf{(CaCE~\cite{goyal2019explaining})} investigates the causal effect made by the presence or absence of a symbolic concept on a network prediction. 
To measure how much a concept can change the classifier's output, they propose operating over the original sample, removing the concept and generating a counterfactual sample.
The CaCE measure is defined as the effect of the presence of the binary concept on the classifier's output. 
To effectively compute it, we must control the generative process of the samples annotated with concepts. 
To achieve this, the authors propose two approaches.
(i) Using a controlled environment, it is possible to intervene in the sample generation process, manually inserting or removing a concept and calculating the exact CaCE value for this intervention. 
(ii) Training a generative model to represent images with and without the concepts to approximate the generative process. 
In the second scenario, a 
Variational Auto-Encoder (VAE) architecture can be used in two different variants, either generating images with one concept or another, or 
generating the counterfactual of an image with/without a certain concept. 
The authors apply their approach to four datasets, comparing CaCE with a non-causal version of CaCE and TCAV~\cite{kim2018interpretability},
showing that CaCE better understands and identifies the concept as having a causal relation with the class prediction. 

\smallskip
Similarly to CaCE, the Causal Proxy Model \textbf{(CPM~\cite{wu2023causal})} provides causal concept-based explanations, this time tailored for Natural Language Processing (NLP) models.
CPM builds upon the CEBaB benchmark~\cite{abraham2022cebab}, a dataset 
studying the presence/absence of a concept on textual data and its causal effect on a model. Starting from a dataset of factual restaurant reviews, it provides counterfactual examples 
in which a concept (food, service, ambiance, etc.) has been modified. 
CPM is first initialized with the weights of the black-box model to explain. For both factual and counterfactual inputs, CPM is then trained to mimic the behaviour of the black-box model. Since counterfactual examples are not provided at test time, CPM is still prompted with the factual sample and information coming from the counterfactual one for counterfactual training. 
CPM explanations 
are compared with TCAV~\cite{kim2018interpretability}, the baseline of CaCE~\cite{goyal2019explaining}, 
and others. The comparison is performed by measuring the distance between the ground-truth (i.e., the actual prediction variation when prompted with a counterfactual sample) and the relative importance of the concept for the predicted class provided by each explainer.  
The results show that both versions of the CPM significantly outperform previous methods in assessing the causality of a concept.

\subsubsection{\nodeconcass{} methods}
\label{sec:post-sup-node-concept-assoc}

\begin{figure}[t]
    \centering
    \includegraphics[scale=0.33]{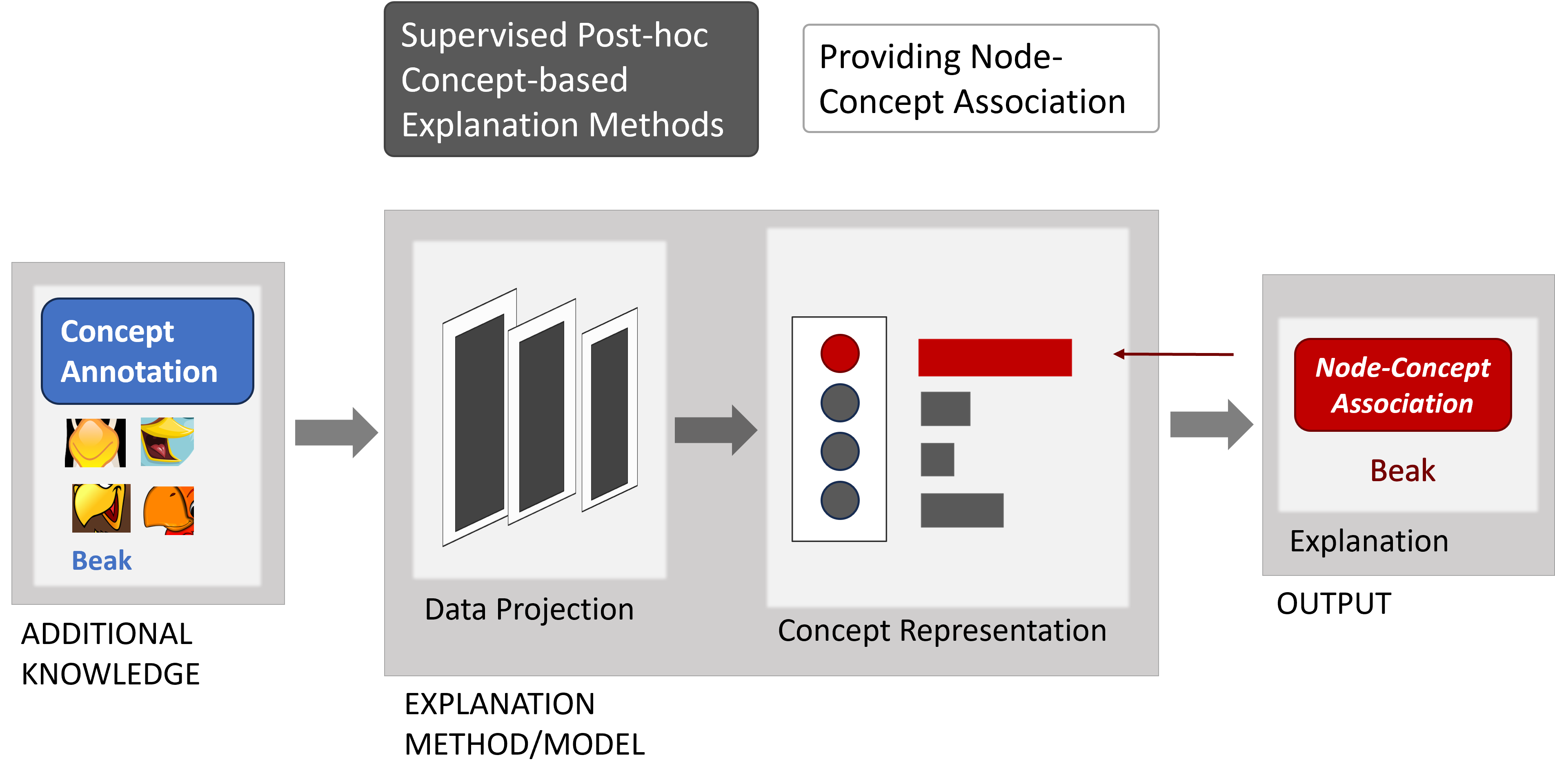}
    \caption{Post-hoc supervised methods providing an explanation in terms of Node-Concept Associations. This class of approaches associates concepts with internal nodes or filters of the network to increase the transparency of the network decision-making process.}
    \label{fig:post_hoc_sup_node}
\end{figure}

Supervised post-hoc methods may also aim to associate symbolic concepts with internal nodes or filters of the neural model. In this case, the idea is to understand where the model has automatically learned some concepts. This allows us to shed some light on the decision process of a given model. As shown in Figure~\ref{fig:post_hoc_sup_node}, these methods supply \nodeconcass{} explanations by checking which nodes activate the most when the network is presented with input image representing the concepts (e.g., the beak) (ND, Net2Vec) or input graphs (GNN-CI). However, this process can also be performed by human annotators analyzing a node's common activation areas (Object Detectors).   

\smallskip
\textbf{Object Detectors~\cite{zhou2015object}} is a precursor to what has been termed concept-based explainability since 2017. 
It aims to shed some light on the learned representations in the hidden layers of a neural network. In particular, the study reveals that hidden neurons automatically learn to recognize some objects when training a network to classify 
scenes. 
More precisely, to understand and quantify the precise semantics learned from each unit, this work uses AMT workers. 
Workers have to uniquely identify the concept 
given the input activation of each unit, as well as define the unit's precision by selecting all the activations that do not correspond to the concept. According to the AMT workers, the results show that a hidden unit automatically learns to recognize a concept with a precision higher than 70\% on average across all hidden layers. 
Interestingly, they show that the types of concepts learned gradually shift from colours and textures in lower layers to objects and object parts in higher ones. 
It can be said that this is the first example of human-supervised annotation of concepts.

\smallskip
Similarly to Object Detectors, Network Dissection \textbf{(ND~\cite{bau2017network})} 
aims to explain single network neurons by associating them with a unique concept.
Rather than employing AMT workers, they propose a new dataset, BRODEN (BROad and DENsely labeled dataset)~\cite{agrawal2014analyzing}, which includes concept labels from diverse data sources. 
To associate each node to a specific concept, the authors propose to compute the Intersection over Unions (IoU) of the bounding boxes representing the concepts and the activation maps of the node over the BRODEN dataset. The node is then associated with the concept with the highest IoU. 
In output, ND reports the units with the highest average IoU and the samples with the highest overlap for each concept.  
For each concept, we can have more than a unit in the network that is associated with it. 
It is essential to notice that Network Dissection's effectiveness relies on the quality of the underlying dataset. If a network unit corresponds to a concept understandable to humans but not present in the given dataset, it will receive a lower interpretability score IoU.
 As a result, they show that individual units may automatically learn to represent high-level semantics objects. 
concluding that the units of a deep representation may be more interpretable than expected. 

\smallskip
The work Network to Vector \textbf{(Net2Vec~\cite{fong2018net2vec})} 
argues that semantic representations of concepts might be distributed across filters. Hence, filters must be studied in conjunction, not individually, as in ND~\cite{agrawal2014analyzing}. 
Net2Vec aligns concepts to filters of a CNN via a two-step approach.
For each concept, Net2Vec first collects filter activations of the network to explain when it is probed with input samples from a dataset annotated with concepts (e.g., BRODEN~\cite{agrawal2014analyzing}).
Then, Net2Vec performs an optimization process to either classify or segment the same concept starting from these activations. The resulting weights tell us how much each node is relevant for predicting the given concept. 
The results showed that combining filters allows to identify better concepts than using a single filter in both tasks, indicating that multiple filters better represent concepts. 
Moreover, filters are not concept-specific, but they can encode multiple ones. 
As a limitation, Net2Vec models the relationship between concepts and filters as linear, potentially failing to capture more complex and non-linear concept alignments.

\smallskip
In Graph Neural Networks Concept-based Interpretability \textbf{(GNN-CI~\cite{xuanyuan2023global})}, the authors extend the idea of node-concept explanations to the challenging graph classification scenario. In particular, they define concepts either as properties of a node in the graph (e.g., number of connected edges > 7) or as the class of the node (being an Oxygen atom) or of the neighbors (next to an Oxygen atom). 
For each neuron, they check whether its activation resembles the presence of any concepts. The resemblance is computed as the IoU between concept presence and neuron activation. Rather than considering a single concept, they allow each neuron to be described as a short composition of concepts, e.g., $\text{number edges} > 7\land \neg \text{is Oxygen}$. 
Furthermore, they produce an explanation of the final class in terms of the concepts as a linear classifier trained only on explained nodes to mimic the original classifier.
Finally, they also propose a concept visualization in terms of a graph concept activation map, a variant of Grad-CAM applied to a graph. 
The experiments show that extracted concepts are meaningful for graph classification. For instance, they reveal the presence of functional groups which are known to be correlated with the graph class. Also, they show that interpretable neuron activations are positively correlated to correct class predictions. 

\subsection{Unsupervised Concept-based Explainability methods}
\label{sec:unsup_concept_exp}
\begin{figure}[t]
    \centering
    \includegraphics[scale=0.33]{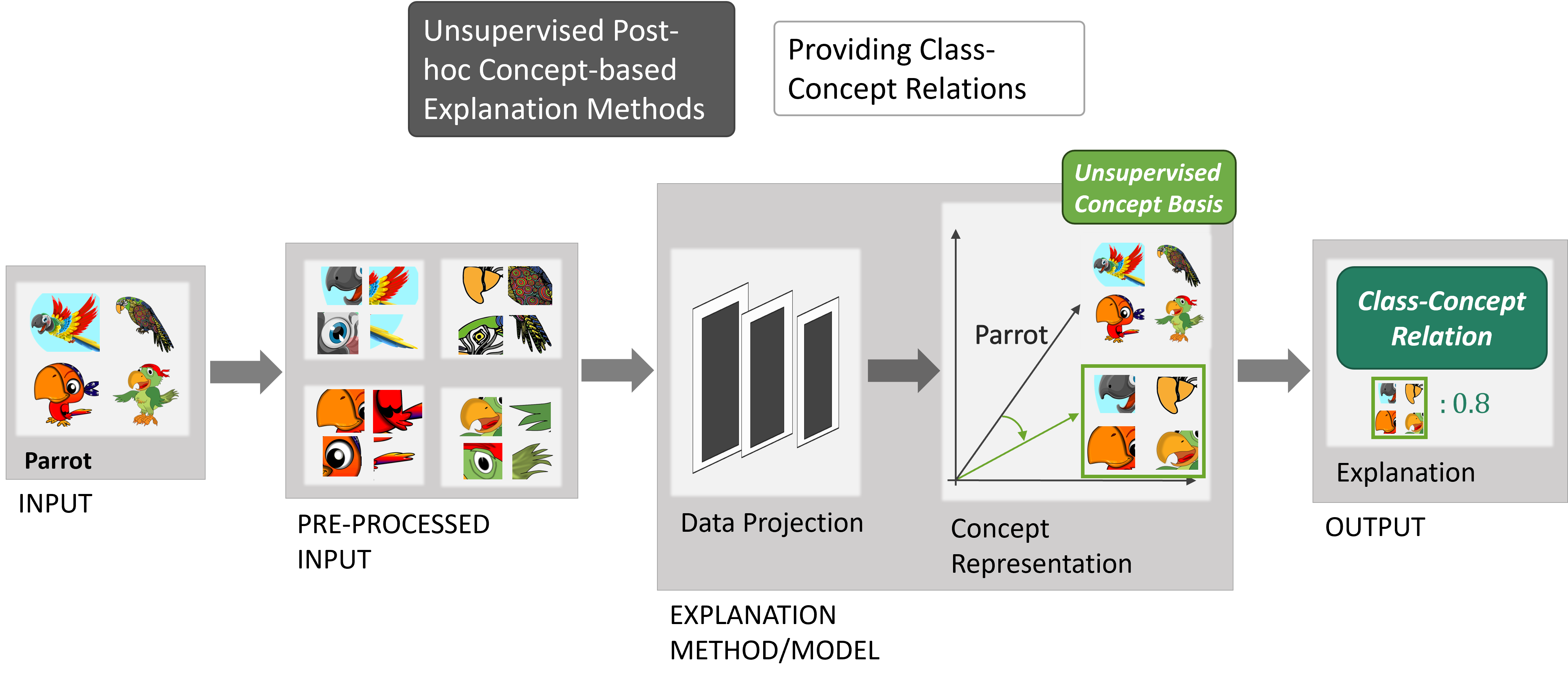}
    \caption{Unsupervised post-hoc methods extract unsupervised concepts in terms of parts of the same input samples. In particular, these methods provide class-concept relations analyzing the representations of these parts with respect to the output class in the latent space of the network.}
    \label{fig:posthoc_unsup_classconc}
\end{figure}

Relying on a given set of supervised concepts is not the only way to assess whether a network is learning higher-level semantics. 
Indeed, parts of the samples may represent practical explanations of a given prediction or of a learned class. 
As shown in Figure~\ref{fig:posthoc_unsup_classconc}, input pre-processing is normally necessary for these methods.
Afterward, unsupervised methods analyze sample representation in the latent space to identify clusters composed of parts of the samples (i.e., \textit{unsupervised concept basis}). 
%
Also, they analyze how these clusters contribute to the final prediction, thereby providing explanations in terms of Class-concept relations.
How the clustering is conducted differentiates the methods: they either employ standard K-means (ACE), Non-Negative Matrix Factorization (ICE), recursion on different layers (CRAFT), subspace clustering (MCD), concept completeness maximization (Completeness-aware), or assuming concept independency, or, finally, working on voxels from videos. 
This category bears similarities to standard XAI methods. However, they differ in not selecting features relying solely on their saliency or internal similarity (as with superpixels) but rather on their capability to represent common patterns existing in other input samples. 


\smallskip
The Automatic Concept-based Explanations \textbf{(ACE~\cite{ghorbani2019towards})} method explains a class within a trained classifier, employing a probing strategy that does not require human supervision.
The process involves first segmenting images belonging to a specific class at multiple resolutions, encompassing textures, object parts, and complete objects, thus capturing a diverse spectrum of concepts.
Subsequently, the projection of image segments in the latent space of the network is clustered across different data.
Outlier sample projections are filtered out since they possibly represent insignificant segments or underrepresented concepts. 
Finally, the T-CAV importance score \cite{kim2018interpretability} is calculated for each group (i.e., a concept) to estimate the average influence of a concept on class prediction. 
The final output comprises the most pertinent concepts based on their T-CAV score. 
As recognized by the authors, however, ACE 
might find meaningless or non-coherent concepts due to segmentation, clustering, or similarity score calculation errors. 
With a human-based evaluation, however, they tested that the samples assigned by ACE to the same concepts are very coherent and can be labeled by a human employing, in general, the same words.

\smallskip
While ACE allows the extraction of some concepts from a learned neural network, there is no guarantee that this set of concepts is sufficient to explain a prediction. In \textbf{Completeness-aware~\cite{yeh2020completeness}}, the authors propose a method for extracting complete concept-based explanations.   
They define a completeness score as the amount to which
concept scores are sufficient for predicting model outcomes. To measure the concept completeness, they propose to assess the accuracy achieved by predicting the class label, given the concept scores. 
With this in mind, they propose a 
concept discovery algorithm is optimized to unveil maximally complete concepts. 
The algorithm also promotes proximity between each concept and its closest neighbouring patches to extract meaningful concepts. 
Furthermore, the authors propose ConceptSHAP, a metric measuring the importance of each concept for a  given class. It modifies SHAP by replacing attribute importance with concept sufficiency in representing the class. 
Using a synthetic dataset with complete concepts, the authors show that the proposed method excels in discovering the complete set of concepts, outperforming compared concept discovery methods, including ACE. 

\smallskip
Invertible Concept-based Explanation \textbf{(ICE~\cite{zhang2021invertible})} builds upon ACE to improve its concept identification and importance assessment algorithm.
ICE replaces K-means with Non-Negative Matrix Factorization (NMF) for extracting concept vectors. 
Using NMF over the feature maps uncovers a concept for each reduced dimension. This allows ICE to employ a classifier on top to approximate the model outputs and yield concept importance as the weights of the linear approximation instead of using T-CAV scores. 
The approximation error (or the completeness of the representation, as defined in \cite{yeh2020completeness}) is much lower when employing the richer NMF concept representation than when employing Principal Component Analysis (PCA) or K-Means, which can also be seen as a form of binary dimensionality reduction. 
Similarly to ACE, 
the authors perform a user study to assess the interpretability of the extracted concepts. They ask participants to assign a few-word description to a group of images belonging to the same extracted concept.
The results show that ICE concepts are more interpretable than those derived from PCA or K-Means.

\smallskip
Concept Recursive Activation FacTorization for Explainability \textbf{(CRAFT~\cite{fel2023craft})} extends the techniques introduced in ACE and ICE for unsupervised concept discovery. 
Also, it offers the possibility of visualizing the unsupervised concepts that contributed to the final prediction. 
In contrast to ACE and ICE approaches, in CRAFT, sub-regions are identified with a random crop 
technique.  
The activations from an intermediate layer of a neural network are calculated for each random crop. These are then factorized using the NMF technique to obtain 
concept representations similar to ICE. 
Selecting the right layer to analyze is critical because concepts can manifest across multiple layers. To address this, CRAFT introduces a recursive approach to identifying concepts and sub-concepts. 
Concept importance for the final prediction is determined using \textit{Sobol} scores derived from sensitivity analysis.
For concept visualization, CRAFT employs a saliency map starting from the concept representation and highlighting the most influential parts in the input image for the concept.
A Human evaluation reports CRAFT's higher utility than standard XAI methods (e.g., Saliency, GradCAM) and ACE in two out of three scenarios. To assess the meaningfulness of extracted concepts, participants were required to detect intruder concepts and identify parent concept vs sub-concept identification to evaluate the effectiveness of the recursive approach.

\smallskip
Multi-dimensional Concept Discovery \textbf{(MCD~\cite{vielhaben2023multi})} introduces multidimensionality in concept extraction by allowing concepts to span on a hyperplane across different convolutional channel directions. This is obtained by means of  
Sparse Subspace Clustering (SSC) for feature vector clustering and a subsequent PCA for cluster basis derivation. On top, a linear layer is employed (similarly to ICE) to mimic the final classification. This allows each sample to define a concept completeness score, quantifying the fraction of the model prediction that can be collectively recovered using the extracted concepts. 
The MCD framework offers two explanations: concept relevance maps, indicating a concept's importance for class predictions, and concept activation maps, for concept visualization. 
In the experiments, the authors show that MCD with an SSC clustering strategy overcomes other concept discovery methods (ICE, ACE) regarding concept faithfulness and conciseness. These are respectively defined as the reduction of the prediction when flipping the most important concepts and as the number of concepts required to reach a certain completeness score. 

\smallskip
The objective of \textbf{DMA \& IMA~\cite{leemann2023are}} 
is to devise a concept discovery method characterized by \textit{identifiability}, extending the idea from PCA and Independent Component Analysis (ICA) to conceptual explanations.
The authors define a method as \textit{identifiable} when it can provably recover all the concepts in a context where the components generating the data are known. 
They propose Disjoint Mechanism Analysis (DMA), which ensures identifiability in compositional data generation where distinct components control distinct pixel groups in an image. DMA guarantees the recoverability 
even if ground truth components are correlated (unlike PCA or ICA). However, it requires components to affect disjoint parts of the input, which should be true as long as the explained encoder is faithful to the generative process generating input data. The second approach, Independent Mechanism Analysis (IMA), 
represents a less constrained version requiring concept vectors 
to be only orthogonal but not disjoint.
The results demonstrate that 
IMA and DMA yield results comparable to PCA or ICA on datasets without correlated generative components and better on the contrary. 
Also, discovered concepts represent ground truth concepts more often than ACE and Completeness-aware.
However, it's worth noticing that DMA and IMA do not provide any of the explanations defined in Section~\ref{sec:expltype}: they only assess which concepts can be extracted from a neural representation. 

\smallskip
Spatial-temporal Concept-based Explanation \textbf{(STCE~\cite{ji2023spatial})} provides concept-based explanations for 3D ConvNets, a domain less explored due to the computational complexity of handling video data. 
STCE directly extends ACE's 2D image strategy to the video domain without neglecting the spatio-temporal dimension. 
The method involves segmenting videos into supervoxels for each class, extracting features with a 3D ConvNet, and clustering similar supervoxels to define unsupervised spatial-temporal concepts.
 Concept importance is determined using T-CAV scores, considering, as negative samples, random videos from non-related datasets.
Similarly to previous works, the quality of the STCE explanation is assessed in terms of faithfulness by measuring the reduction of accuracy of the explained model when removing concepts associated with high T-CAV scores. 
For qualitative analysis, the authors present video frames illustrating detected concepts, particularly by highlighting video frames that feature the most and least important concepts for a given classification 

\subsection{Discussion}
\label{sec:posthoc_issues}
In the following, we report the critics that have been moved to the concept-based methods reviewed in this section and how, in particular, they affect their performance.  

\myparagraphbold{Do standard networks learn concepts?}
Several works in literature have disputed it \cite{koh2020concept, chen2020concept}
whether standard models actually learn symbolic concepts when trained to accomplish a different task. In particular, \cite{koh2020concept} tested the concept accuracy of a linear probe. We remember that a linear probe is a linear layer trained to predict the concepts in the latent space of an already trained model, as proposed in \cite{kim2018interpretability}. They found that linear probes achieve a lower concept accuracy with respect to a model trained to predict the concepts (91\% vs 97\% on the CUB dataset~\cite{} and 0.68 vs. 0.53 error on the OAI dataset). 
This result confirms that post-hoc concept-based explainability methods cannot ensure that the model actually learns symbolic concepts, as introduced at the beginning of this section. On the other side, they help to understand what the network is learning: the still high accuracy of linear probes implies that the model learned many concepts. 


\myparagraphbold{Different probe datasets provide different explanations}
In \cite{ramaswamy2023overlooked}, the authors tested the importance of the dataset choice in extracting the explanations.  
They have shown that different probe datasets lead to different explanations even when explaining the same model using the same interpretability method. This issue arises for methods explaining both node-concept associations and class-concept relations. As an example of the first type, they show that ND label neurons differently according to the dataset employed (ADE20k or Pascal): while in some cases, the concepts are still similar (e.g., \textit{plant} vs \textit{potted-plant}, \textit{computer} vs \textit{tv}), in other cases are profoundly different (e.g., \textit{chair} vs \textit{horse}, \textit{tent} vs \textit{bus}).
As an example of the second type, they show that the T-CAV concept activation vectors extracted from different datasets are projected in different directions. To numerically assess this, they compute the cosine similarity between the vectors representing the same concept but extracted from ADE20k and Pascal, and they show that while for same concepts the similarity score is quite high (\textit{ceiling} 0.27), for others it is very low (\textit{bag} 0.01, \textit{rock} -0.02), with a cosine similarity of common concept vector lower than 0.1 on average.
As we also highlighted in Section \ref{sec:categorization}, they suggest that the probe dataset should be chosen carefully and specifically recommend probe datasets that closely resemble the data distribution of the dataset on which the model was originally trained.

\myparagraphbold{Post-hoc concept-based explanation methods are vulnerable to adversarial attack}
In \cite{brown2023making}, Concept-based post-hoc explainability methods, much like their standard XAI counterparts \cite{kindermans2019unreliability,ghorbani2019interpretation, slack2020fooling,adebayo2021post}, are susceptible to adversarial perturbations. 
This study focuses primarily on supervised methods that explain class-concept relationships using linear probes, such as TCAV. The underlying idea is that a sample can be crafted to change the interpretation while maintaining the same classification. In particular, they show that an attack can induce a positive interpretation with respect to a concept, such as making a \textit{Corgis} important for the classification of a \textit{Honeycomb}. At the same time, they also provide examples of negative interpretation, such as making \textit{Stripes} not important for classifying a \textit{Zebra}. Notably, the authors also demonstrate that these attacks maintain some level of effectiveness even in black-box scenarios, i.e., when the adversarial samples are created for a different model to the one they are tested on, which is a more realistic scenario.   

\section{\conceptmodel{}} 
\label{sec:concept_model}

\conceptmodel{} introduces a deviation from standard neural network training practices. They explicitly represent concepts within the same neural network architecture. We define them as explainable-by-design since they inherently provide Node-concept associations (N-CA) (except for some cases). 
They typically employ a dedicated hidden layer to predict concept scores to this aim. These scores represent numerical values quantifying the relevance or presence of specific concepts in a given input sample and condition the model's output. The same concept scores can also be studied in relation to the output classes to define Class-Concept Relations (C-CR) and can be visualized through Concept-Visualization (C-Viz).  

How the intermediate representation is defined varies according to the concept annotations and, consequently, the concept types. As shown in Table~\ref{tab:summary:bydesign1}, 
Concept-based models either employ a dataset with annotated concepts (Supervised, Section~\ref{sec:sup_concept_model}), without concept annotations (Unsupervised, Section~\ref{sec:unsup_concept_model}), with annotations for few concepts only (Hybrid, Section~\ref{sec:sup_unsup_cbm}) or they generate the supervision by means of an external model (Generative, Section~\ref{sec:generative}). 
Furthermore, if they employ annotated symbolic concepts (Symb.), they differ according to whether they employ them during training (Joint) or afterward to instill them in a trained network (Instil). If they automatically extract concepts, they differ for the type of concepts extracted, either clusters (Uns. Basis) or prototypes (Proto.). If they adopt a hybrid approach, they generally employ both symbolic and unsupervised basis concepts. Finally, if they generate concept annotations, these are normally Textual concepts.   
Concept-based models have been mostly developed to solve classification tasks (CLF), with one also performing regression (REG) and another clustering (CLU). Similar to the post-hoc method, in this case, the network's backbone is mostly a CNN working on images (IMG), possibly reconstructing the data with an auto-encoder (AE). Some works employ a fully-connected network working on tabular data (TAB), GNN on graphs, and transformer (TRANS) working on both images and texts (TXT).
To summarize the key aspects of concept-based models, we now discuss their main advantages and disadvantages compared to post-hoc methods.

\begin{table}[t]
\caption{Explainable by-Design Concept-based Models. We characterize the approaches based on the following dimensions. 
    Full category description in Section~\ref{sec:categorization}.}
    \label{tab:summary:bydesign1}
\resizebox{\textwidth}{!}{
\begin{tabular}{ll|l|lccccccc}
&  & \textbf{Method} & \multicolumn{1}{c}{\textbf{\begin{tabular}[c]{@{}c@{}}Concept \\ Employ.\end{tabular}}} & \multicolumn{1}{c}{\textbf{\begin{tabular}[c]{@{}c@{}}Concept \\ type\end{tabular}}} & \textbf{Scope} & \multicolumn{1}{c}{\textbf{\begin{tabular}[c]{@{}c@{}}Expl\\ Type\end{tabular}}} & \textbf{\begin{tabular}[c]{@{}c@{}}Data \\ type\end{tabular}} & \textbf{Loss} & \textbf{Task} & \textbf{\begin{tabular}[c]{@{}c@{}}Network\\ Type\end{tabular}} \\ \hline

\multirow{20}{*}{\rotatebox[origin=c]{90}{\textbf{Concept Annotation}\hspace{1cm}}}
& \multirow{9}{*}{\rotatebox{90}{Supervised}} 
& CBM \cite{koh2020concept}                             & \multirow{5}{*}{Joint} & \multirow{5}{*}{Symb.}& N-CA, C-CR    & GL        & IMG & \vmark& CLF, REG & CNN \\
&  & LEN \cite{barbiero2022entropy, ciravegna2023logic} &  & & N-CA, C-CR    & LO \& GL  & TAB, IMG, TXT & \vmark& CLF, CLU & FCN, CNN \\
&  & CEM\cite{zarlenga2022concept}                      &  & & N-CA          & GL        & TAB, IMG & \xmark & CLF & FCN, CNN \\
& & ProbCBM \cite{kim2023probabilistic}                 &  & & N-CA, C-CR    & GL        & IMG & \xmark/$ \approx$ & CLF & CNN \\
&  & DCR\cite{barbiero2023interpretable}                &  & & N-CA, C-CR    & LO        & TAB, IMG, GRA & \xmark & CLF & FCN, CNN, GNN \\
\cline{3-11}
&  & CW  \cite{chen2020concept}                         & \multirow{4}{*}{Instill} & \multirow{4}{*}{Symb.}           & N-CA          & GL        & IMG    & \vmark& CLF & CNN \\
&  & CME   \cite{kazhdan2020now}                        &  &  & N-CA, C-CR          & GL        & IMG    & \vmark & CLF & CNN \\
&  & PCBM   \cite{yuksekgonul2022posthoc}               &  &  & N-CA, C-CR    & GL        & IMG & \vmark /$\approx$ & CLF & CNN \\
&  & CT   \cite{rigotti2022attentionbased}              &  &  & N-CA, C-CR    & LO \& GL  & IMG & \xmark & CLF & TRANSF \\ 
\cline{2-11}
& \multirow{11}{*}{\rotatebox{90}{Unsupervised}} 
& Interp. CNN \cite{zhang2018interpretable}             & \multirow{4}{*}{-} & \multirow{5}{*}{Uns. Basis} & N-CA, C-Viz       & GL        & IMG & \vmark/$ \approx$  & CLF & CNN \\ 
& & SENN   \cite{alvarez2018towards}                    &  & & N-CA, C-CR        & LO        & IMG & \xmark/$ \approx$ & CLF & CNN + AE \\
&  & BotCL  \cite{wang2023learning}                     &  & & N-CA, C-CR              & LO        & IMG & \xmark/$ \approx$ & CLF & CNN + AE \\
&  & SelfExplain   \cite{rajagopal2021selfexplain}      &  & & C-CR              & LO \& GL  & TXT & \xmark  & CLF & TRANSF \\
\cline{3-11}
&  & PrototypeDL \cite{li2018deep}                      & \multirow{6}{*}{-} & \multirow{6}{*}{Proto.}                & N-CA, C-CR, C-Viz & GL        & IMG & \xmark & CLF & AE \\
&  & ProtoPNet \cite{chen2019looks}                     &  &  & N-CA, C-CR, C-Viz & GL        & IMG & \vmark& CLF & CNN \\ 
&  & ProtoPool \cite{rymarczyk2022interpretable}        &  &  & N-CA, C-CR, C-Viz & GL        & IMG & \xmark & CLF & CNN \\ 
&   & Def. ProtoPNet~\cite{donnelly2022deformable}      &  &  & N-CA, C-CR, C-Viz & GL        & IMG & \xmark & CLF & CNN \\ 
& & HPNet \cite{hase2019interpretable}                  &  &  & N-CA, C-CR, C-Viz & GL        & IMG &  \xmark & CLF & CNN \\ 
&  & ProtoPShare   \cite{rymarczyk2021protopshare}      &  &  & N-CA, C-CR, C-Viz & GL        & IMG  & \xmark  & CLF & CNN \\ 
&  & ProtoPDebug \cite{bontempelli2023conceptlevel}     &  &  & C-Viz             & GL        & IMG & -  & CLF  & CNN  \\
\cline{2-11}
& \multirow{3}{*}{\rotatebox{90}{Hybrid}} 
& CBM-AUC \cite{sawada2022concept}                      &  \multirow{3}{*}{-}& \multirow{3}{*}{\makecell{Uns. Basis,\\ Symb.}}     & N-CA, C-CR    & GL        & IMG + VID & \xmark/$ \approx$ & CLF & CNN \\
&  & Ante-hoc \cite{sarkar2022framework}                &  &  & N-CA          & LO \& GL  & IMG & \vmark& CLF &  CNN + AE\\ 
&  & GlanceNets   \cite{marconato2022glancenets}        &  &  & N-CA, C-CR    & GL        & IMG & \xmark & CLF & CNN + VAE \\
\cline{2-11}
& \multirow{2}{*}{\rotatebox{90}{Gen.}} 
& LaBO \cite{yang2023language}                          &  \multirow{2}{*}{-} & \multirow{2}{*}{Textual} & C-CR & GL          & IMG & \xmark/$ \approx$ & CLF & CNN+LLM \\
&  & Label-free CBM   \cite{oikarinen2023label}         &  &  &  C-CR & LO \& GL    & IMG & \vmark& CLF & CNN+LLM \\ \hline
\end{tabular}
}
\end{table}


\myparagraphbold{Advantages}
Concept-based Models offer the advantage of ensuring that the network learns a set of concepts explicitly. 
Furthermore, 
a domain expert can modify the predicted values for specific concepts and observe how the model's output changes in response. This process is referred to as \textit{concept intervention}, and it enables the generation of counterfactual explanations. 

\myparagraphbold{Disadvantages}
On the other hand, these methods can only be employed when there is flexibility to train a model from scratch, possibly tailoring it to the specific task at hand.  Furthermore, in simpler solutions, the predictive accuracy of concept-based models may be lower than that of standard black-box models. Further issues and challenges of concept-based models will be analyzed at the end of the section (Section \ref{sec:cbm_issues}).


\subsection{Supervised Concept-based Models}
\label{sec:sup_concept_model}
Supervised Concept-based Models employ a dataset annotated with symbolic concepts (like attributes of the output classes) to supervise an intermediate layer explicitly representing the concepts. 
When these annotations are readily available within the same training dataset, a \textit{joint concept training} strategy can be employed (Section \ref{sec:sup_joint_train}). 
Otherwise, they can also be embedded in the model through separate training on an external dataset dedicated solely to concept learning to perform a \textit{concept instillation} (Section \ref{sec:sup_concept_instill}).

\subsubsection{Joint concept training} \label{sec:sup_joint_train}
\begin{figure}[t]
    \centering
    \includegraphics[scale=0.33]{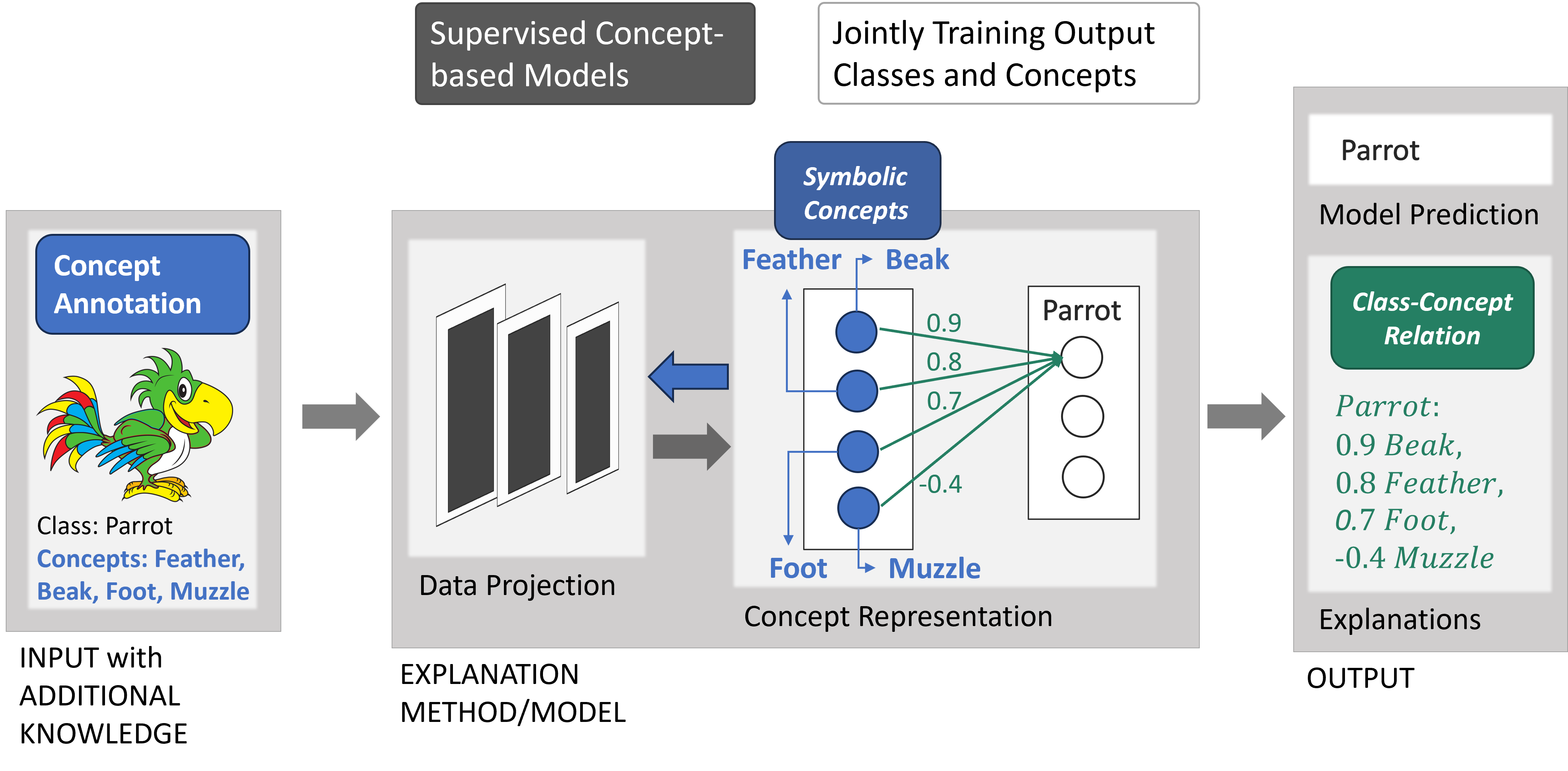}
    \caption{Supervised Concept-based Models jointly training output classes and concept representation on a dataset where both class and concept annotation is provided.}
    \label{fig:cbm_sup_jt}
\end{figure}

 As shown in Figure~\ref{fig:cbm_sup_jt}, supervised concept-based models employ concept annotations to supervise an intermediate representation and increase the transparency of the model. 
 Unlike methods performing concept instillation, the training is performed \textit{jointly} for the target task (e.g., image classification) and for concept representation.
 This requires employing a training dataset that includes annotations about both the main classes (e.g., Parrot) and the related attributes (e.g., Feather, Beak, Foot, Muzzle).  
 Concepts can be represented by means of a single neuron (CBM, LEN) or through an embedding (CEM, ProbCBM, DCR), which avoids performance loss with respect to a black-box model. As shown in the picture, these methods can also analyze the relations between concepts and classes through concept importance (CBM, ProbCBM) or logic relations (LEN, DCR).    

\smallskip
The first work proposing a joint task-concept training strategy is attributed to
Concept-Bottleneck Model \textbf{(CBM~\cite{koh2020concept})}, where 
the authors modify the typical end-to-end neural architecture mapping from an input to a target by introducing an intermediate \textit{concept bottleneck} layer. 
This layer has neurons dedicated to learning and predicting a set of human-specified concepts from input features. A task function then predicts the final target, relying only on concept prediction (concept scores). 
Specifically, 
a double loss is optimized, penalizing wrong predictions in both the output and concept bottleneck layers. 
CBM design enhances the model's transparency and enables interaction with the model itself. 
A domain expert may intervene in the predicted concepts, allowing potential adjustments to the model's predictions. 
This process is referred to as \textit{concept intervention}. 
Concept interventions also facilitate the extraction of counterfactual explanations by manipulating the concept predictions and observing the model's response on the counterfactual concept representation. 
CBMs, however, are affected by two main limitations. 
i) CBMs provide class-concept relations only when employing a single layer to model the task function. In this case, the importance of a concept corresponds to the weight connecting the concept to the final class, as shown in Figure \ref{fig:cbm_sup_jt}. 
ii) The generalization capability of CBM is lower than standard end-to-end models. This is mainly due to the bottleneck layer where each concept is represented by a single neuron~\cite{mahinpei2021promises}. This limit becomes more severe in contexts with few concepts available and when employing a shallow task function.

\smallskip
In a series of works, the authors address the first issue by means of a framework named Logic Explained Networks \textbf{(LENs ~\cite{barbiero2022entropy, ciravegna2023logic})}. 
LENs improve the explainability of CBMs. These models, indeed, are trained to provide explanations in terms of First-Order-Logic (FOL) formulas associating task functions with the concept having the highest influence on it. To achieve this goal, they impose architectural constraints enforcing sparsity in the weights of the task function. As an example, on a bird classification task, a LEN may provide explanations of the type $\text{billShapeHooked} \land \text{mediumSize} \land \neg \text{throatColorWhite} \Leftrightarrow \text{blackFootedAlbatross}$. 
While most of the other works focus on vision tasks, LENs can be applied to tabular data and textual data \cite{jain2022extending} without a concept extractor, i.e., without requiring a module explicitly mapping the non-interpretable input data to a concept space. LEN can consider the same input data as concepts when they are associated with human-understandable symbols. 
Furthermore, LEN can solve clustering tasks 
and provide logic explanations about cluster assignment in terms of frequent concept relations~\cite{ciravegna2023logic}.
Finally, logic explanations can be used to defend a LEN from an adversarial attack by checking the consistency of the test-time predictions with the explanations extracted at training time.

Concept Embedding Models \textbf{(CEM~\cite{zarlenga2022concept})} improve CBM performance by using an entire set of neurons to represent a concept (concept embedding). CEM obtains the same classification accuracy of black-box neural networks while retaining the possibility to employ and interact with a concept-based model. Unlike CBM, CEM performance does not decrease when the number of employed concepts decreases, showing a very high concept efficiency.
A naive solution to the performance loss issue of CBM is adding a separate set of unsupervised neurons \cite{mahinpei2021promises}, encoding information about the task function. 
However, as shown in CEM, this hybrid solution loses the possibility of interacting with the network. 
According to CEM authors, in the hybrid configuration, the task functions only rely on unsupervised neurons, nullifying the effects of concept interventions on the task function. 
In CEM, the authors also introduce the Concept Alignment Score to measure the quality of a concept representation in describing a set of concepts. More precisely, it measures the entropy reduction in the concept space when a concept representation is provided. 
They show that CEM concept alignment to the real concept distribution is close to the one provided by CBM and much higher than the hybrid solution.

\smallskip
When concepts used to make the final predictions are not visible in the input sample, a deterministic prediction of the concepts may damage 
both the final task prediction and the explanation based on the concepts. 
To solve this problem, the Probabilistic Concept Bottleneck Model \textbf{(ProbCBM~\cite{kim2023probabilistic})} employs probabilistic concepts to estimate uncertainty in both concept and class predictions. 
Similarly to CEM, ProbCBM also employs concept embedding. However, given an input, ProbCBM outputs mean, and variance vectors 
for each predicted concept. Concepts are then represented as multivariate normal distributions. The probability of 
each concept is calculated via Monte-Carlo estimation by sampling from the distribution and computing the mean distance from the true and false concept vectors. Class predictions are similarly computed, generating a class embedding from the concept embeddings. 
The uncertainty quantification in ProbCBM is done by calculating the determinant of the covariance matrix, which represents the volume of the probabilistic distribution. 
ProbCBM's performance is compared with CBM~\cite{koh2020concept} and CEM~\cite{zarlenga2022concept}. ProbCBM outperforms CBM in classification accuracy but falls short of CEM. Regarding concept prediction accuracy, it is on par with CBM and better than CEM, demonstrating that integrating uncertainty through probabilistic embedding doesn't compromise performance. 

\smallskip
Even though CEM and Prob-CBM enhance the performance of standard CBMs, 
they worsen the underlying explainability. When employing embedded concepts, even with the use of a shallow network, the interpretability of the task function is compromised since the dimensions within a conceptual embedding lack a direct association with a symbolic meaning. 
In response to this challenge, Deep Concept Reasoner \textbf{(DCR~\cite{barbiero2023interpretable})} is introduced  
as an interpretable concept embedding model. 
Starting from concept embeddings, DCR 
learn differentiable modules which output for each class and each sample a different fuzzy rule. These rules are then executed, taking into account the concept activations. 
As a result, the model outputs an interpretable prediction regarding the concepts, even though only locally. 
DCR achieves task accuracies on par with CEM and higher than CBM. The authors also show that in contexts where ground truths rule are available, DCR can recover them. 
Compared with LIME and LEN, DCR facilitates the generation of counterfactual examples, and extracted explanations are less sensitive to input perturbations. 
Finally, they showcase that DCR can also work on graphs employing GNN as a backbone. In this case, they extract concepts post-hoc from a trained GNN by means of a Graph-Concept Explainer~\cite{magister2021gcexplainer}, a variant of ACE for graph structure. 

\subsubsection{Concept Instillation} \label{sec:sup_concept_instill}
\begin{figure}[t]
    \centering
    \includegraphics[scale=0.33]{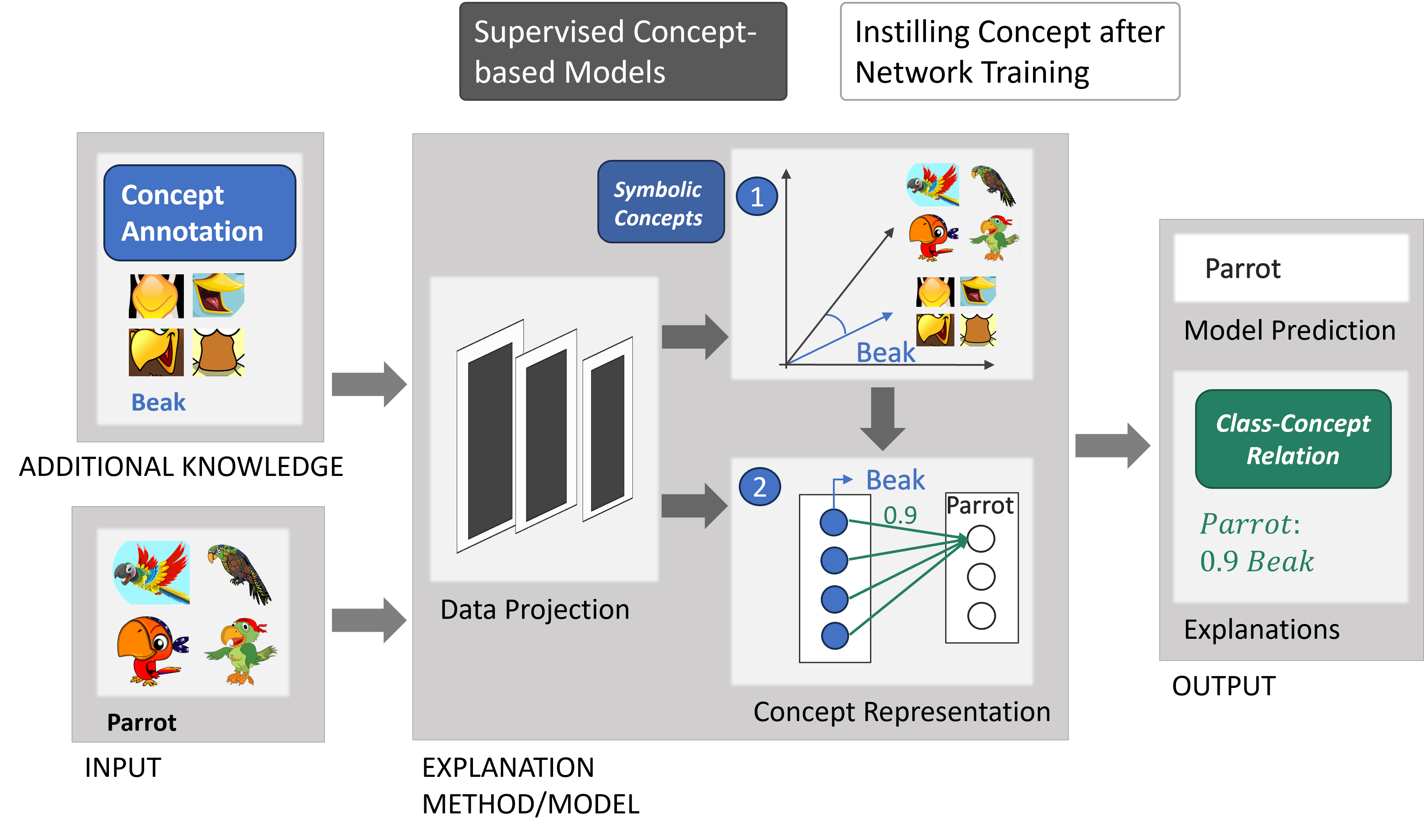}
    \caption{Supervised Concept-based Models instilling concept in the network after a first end-to-end network training. Concept representation is either extracted from the network itself with post-hoc C-XAI techniques or from an external source (e.g., a knowledge graph).}
    \label{fig:cbm_sup_ci}
\end{figure}

This class of approaches considers the case where concept information is not available in the training set, but it is available from an external data source, such as a separate supervised dataset (CW, CT), a semi-supervised one (CME), or a knowledge graph (PCBM).  
As shown in Figure~\ref{fig:cbm_sup_ci}, the process is composed of two steps: 1) a black-box network is standardly trained end-to-end; 2) through \textit{Concept Instillation}, a given layer of the network is modified to represent concepts. At test time, this process still allows the employment of a concept-based model. Similarly to jointly trained models, concepts can be represented by means of single nodes (CW, CME, PCBM) or through concept embeddings (CT). They also provide class-concept relations if they replace the classifier head with an interpretable task predictor (CME, PCBM, CT).

\smallskip
Concept Whitening \textbf{(CW~\cite{chen2020concept})} modifies a hidden layer of a neural network after it has been trained in such a way that it also predicts a given set of concepts. 
Usually, CW can be used to replace the Batch Normalization (BN) layer in a CNN because it combines batch whitening (decorrelating and normalizing each dimension) with an extra step involving a rotation matrix to align the concepts with the axes. 
The underlying idea is that axes of the latent space should align with concepts, allowing each point within the space to correspond to and be interpreted through known concepts. 
Further training on the standard classes is required to avoid loss in the final task performance.
An interesting aspect of CW is its capability to determine how a network captures a concept across different layers, as it can be applied after any layer.  
The authors show that a network produces more abstract and fundamental representations in the lower layers. 
The authors compute the concept purity as the 1-vs-all AUC score on the test concept predictions to evaluate interpretability. 
A comparative analysis against concept-based post-hoc methods (TCAV, IBD) 
indicate that the concepts acquired with CW exhibit higher purity. 
This result is attributed to the inherent orthogonality of CW conceptual representations and the impact of the entire model fine-tuning process.


\smallskip
Similarly to CW, the Concept-based Model Extraction \textbf{(CME~\cite{kazhdan2020now}) }also proposes to derive an interpretable model from a black-box.  
CME trains a concept extractor starting from the hidden representation of a given DNN.
To reduce the concept annotation effort, CME employs only a small annotated dataset and trains the concept extractors in a semi-supervised fashion. Interestingly, the authors repeat the extraction process of each concept for different hidden layers, retaining the one providing the higher concept accuracy. 
The predicted concept scores are employed as input for an interpretable task predictor, which is trained to mimic the task prediction of the original model.  
For the task function, the authors employ Decision Trees and Logistic Regression. 
Since they directly work on concepts, they both allow concept intervention at test time.
CME's performance is compared with Net2Vec~\cite{fong2018net2vec} and CBM~\cite{koh2020concept} frameworks. CME significantly outperforms Net2Vec in task accuracy while remaining comparable to CBM.
The authors also introduce a new metric, the MisPrediction Overlap (MPO), 
to compute the portion of samples with at least \textit{m} concepts inaccurately predicted. 
Based on MPO, CME performs similarly to Net2Vec but falls behind CBM.

\smallskip
Post-hoc Concept Bottleneck Models \textbf{(PCBM~\cite{yuksekgonul2022posthoc})} implement two modalities to extract concepts: (i) from other datasets, similarly to CW~\cite{chen2020concept}, and (ii) from a knowledge graph and then encoded in the network by means of a textual encoder. 
In the first scenario, Concept Activation Vectors (CAVs) are employed to learn the concept representations in the latent space of a model pretrained on the final task. 
In the latter scenario, the authors employ an open-access 
knowledge graph (e.g., ConceptNet~\cite{speer2017conceptnet}), to gather concepts in relation 
with the queried classes. 
They then obtain a vectorial representation of the concepts by means of a text encoder. 
In both scenarios, the concepts are projected in the latent space of a pretrained model. The projections then train an interpretable task predictor, providing class-concept importance. 
In addition to concept intervention, PCBMs also allow for \emph{global model edits}, 
global modifications, which enables the model 
to adapt its behaviour to novel scenarios, including shifts in data distributions. 
Similarly to~\cite{mahinpei2021promises}, the authors propose to employ an unsupervised set of neurons to improve network scaling. 
This time, however, they propose to employ a residual fitting, such that the hybrid prediction still has to rely on the interpretable one.
With respect to the original model, PCBM performances are slightly worse, but still comparable. 

\smallskip
ConceptTransformer \textbf{(CT~\cite{rigotti2022attentionbased})} substitutes the task classifier of a DNN 
with an attention module enriched with concept embeddings. Instead of being extracted from a knowledge base as in PCBM, concept embeddings in CT come from a transformer model trained on a dataset annotated with user-defined symbolic concepts.
The output of the attention module represents concept scores. A linear layer is employed on top to predict the final classes. 
CT provides \classconcrel{} in terms of the attention scores between the output class and the concepts embeddings. 
However, CT does not provide \nodeconcass{} since it does not explicitly represent concepts by means of a set of nodes but only with concept embeddings. 
The authors claim that their model is faithful by-design since it employs a linear layer from concepts to classes, similar to CBM. 
To obtain more plausible explanations, they also propose supervising attention scores so that concept-output relationships align with domain knowledge. 
CT classification performance matches end-to-end black box models. 
Leveraging the concepts with CT seems to even boost the performance in contexts where the set of concepts is well-defined, and the relationships with the final classes are known (as in the case of CUB-200-2011 dataset \cite{welinder2010caltech}). 

\subsection{Unsupervised Concept-based Models}
\label{sec:unsup_concept_model}
Unsupervised concept models (Section \ref{sec:unsup_concept_model}), modify the internal representation of a network to autonomously extract concepts without explicitly associating them with pre-defined symbols. 
In some cases, these models jointly train an intermediate layer with an unsupervised learning technique to extract an \textit{unsupervised concept basis}, i.e., a clusterized representation identifying groups of similar samples (Section \ref{sec:unsup_concept_basis}). 
Otherwise, they explicitly require the network to represent in the layer weights \textit{prototypes-concepts},  (parts of) frequently seen input samples (Section \ref{sec:unsup_prototype}).

\subsubsection{Unsupervised Concept Basis}
\label{sec:unsup_concept_basis}
\begin{figure}[t]
    \centering
    \includegraphics[scale=0.33]{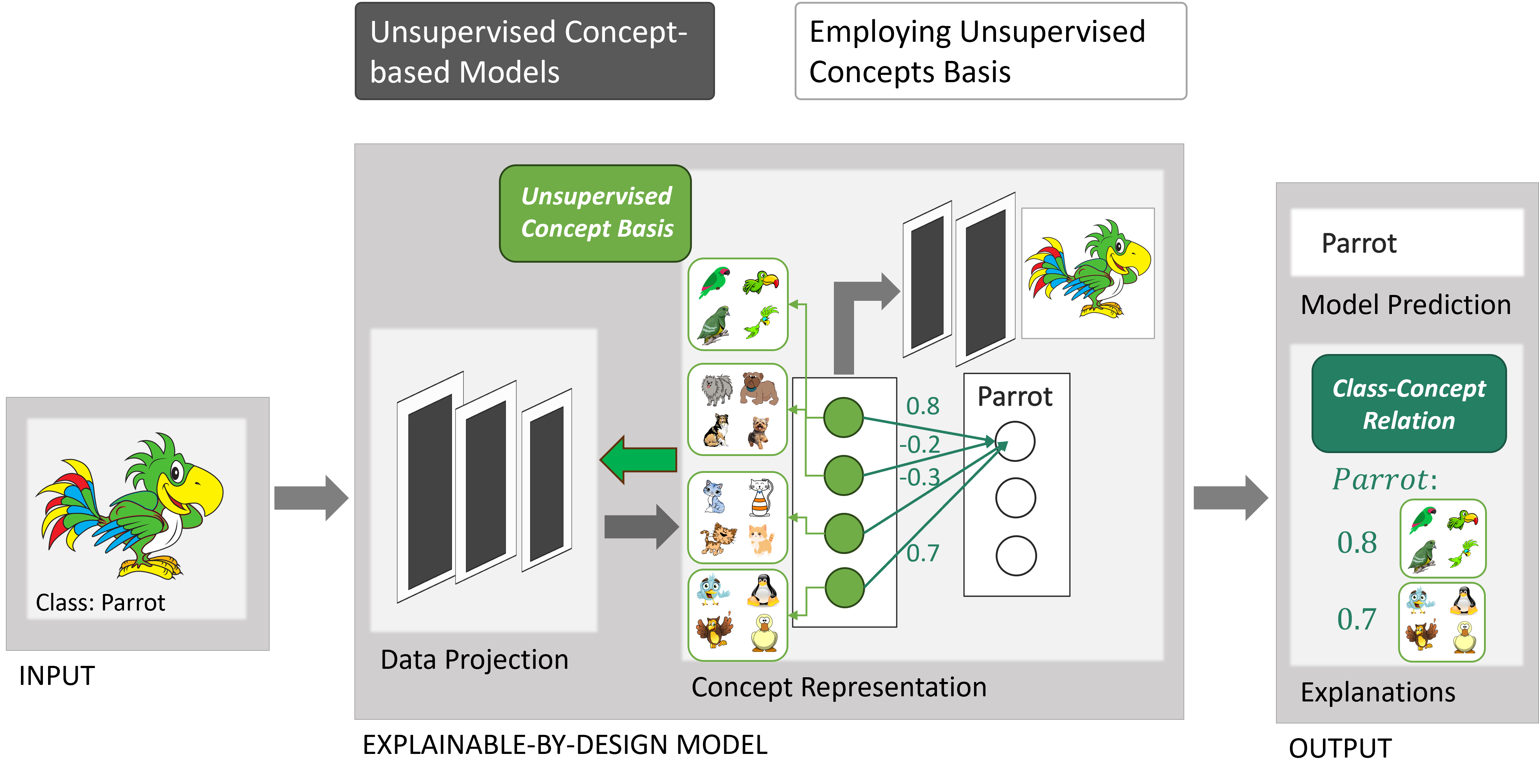}
    \caption{Unsupervised concept-based model extracting unsupervised concept basis. How the concepts are extracted varies between the models. Here, we report the case in which a decoder branch is employed over the concept to reconstruct the input and extract meaningful concepts (SENN, BotCL).}
    \label{fig:cbm_unsup_basis}
\end{figure}

The idea behind these approaches is to learn unsupervised disentangled representations in the latent space of a model. These representations must group samples with some characteristics, such as a generative factor. This class of approaches takes inspiration and shares ideas with the disentangled representation learning field~\cite{bengio2013representation}.
To extract unsupervised concepts, these models are jointly trained to reconstruct the input by means of a decoder branch (SENN, BotCL, and reported in Figure~\ref{fig:cbm_unsup_basis}) to maximize the mutual information (Interpretable CNN) or via self-supervised loss over textual data (SelfExplain). When these models employ an interpretable task predictor over the concepts, they also provide class-concept relations (SENN, BotCL).

\smallskip
In \textbf{Interpretable CNN~\cite{zhang2018interpretable}}, convolutional filters act as unsupervised concepts and are required to represent different object parts. The assignment occurs during the learning process without employing any concept annotation.
The authors devise an unsupervised loss function to achieve this, maximizing the mutual information between the image and the filter activations. 
Also, the loss promotes low entropy in both the inter-category activations and the spatial distributions of neural activations.
The idea is that filters should create a clusterized representation of the concepts. Each filter should encode a distinct concept associated with a single object category, possibly not repeated in different image regions. 
In the evaluation, the authors measure the IoU score with ground-truth object bounding boxes, similar to ND. 
They also measure the part location stability, 
assessing the consistency in how a filter represents the same object part across various objects. 
The experimental findings show that Interpretable CNNs exhibit superior interpretability and location stability compared to standard CNNs. Regarding classification accuracy, Interpretable CNNs are on par with simple CNNs. 

\smallskip
Self-explanatory neural network \textbf{(SENN~\cite{alvarez2018towards})}, similarly to Interpretable CNN, also employs an unsupervised approach to create an explainable-by-design network, this time based on an auto-encoder architecture. 
SENN first employs a concept encoder to derive a clusterized representation of its features from the input. 
Second, it employs an input-dependent parametrizer that generates class-concept relevance scores. 
The output is given by the linear combination of these basis concepts and their relevance scores. 
The degree of interpretability of SENN is intrinsically tied to the interpretability of the underlying concepts. 
SENN promotes concept diversity to this aim: inputs should be representable using a few non-overlapping concepts. This is achieved by making the concept encoder sparse. 
Also, to visualize the identified concepts, SENN 
provides a small set of training examples that maximally activate it.
The authors show that SENN explanations compared to standard XAI techniques (i.e., LIME \cite{ribeiro2016should}, SHAP\cite{lundberg2017unified}) result in more:
i) explicit, SENN explanations are more immediate and understandable;
ii) faithful, SENN importance scores are assigned to features indicative of ground-truth importance;
iii) stable, since similar examples yield similar explanations for SENN. 
SENN classification accuracy is slightly lower or on par with standard CNN according to the experimented scenarios. 

%
\smallskip
Bottleneck Concept Learner \textbf{(BotCL~\cite{wang2023learning}) }is based upon and improves the idea of SENN for extracting an unsupervised concept basis. 
In BotCL, the features extracted from the convolutional encoder are fed into a slot attention-based mechanism~\cite{li2021scouter} predicting for concept scores. A set of vectors learnt by backpropagation represent the concept embeddings and are used as keys of the attention mechanism. 
%
A linear layer working over concept scores is employed as a task classifier, and it captures class-concept relationships. 
%
BotCL employs the reconstruction loss of SENN or a contrastive loss to learn meaningful concepts.
The latter enforces similarity among concept activations belonging to the same class and dissimilarity otherwise. 
%
Also, BotCL employs two regularizations to facilitate training and ensure consistency in the learned concepts.
The first is a consistency loss, enforcing similarity between the concept features that activate the concept in different images. The second is a mutual distinctiveness loss, requiring different concepts to cover different visual characteristics.
BotCL task performance, on average, is higher than SENN, ProtoPNet, and Completeness-aware, particularly when employing the contrastive loss. 
BotCL extracted concepts, instead, are compared against ACE and Completeness-aware on a synthetic dataset annotated with concepts. They result in being good in terms of concept purity and efficiency and reconstruction error, particularly when employing many concepts. 
%
%

\smallskip
\textbf{SelfExplain~\cite{rajagopal2021selfexplain} }is a self-explaining model, providing local explanations within text classifiers as \classconcrel{}. It extracts concepts in an unsupervised way directly from the input text. In particular, the concepts are here represented as the non-terminal leaves of the semantic tree parsing the text. 
SelfExplain incorporates two different layers into a neural text classifier to explain the predictions. 
Specifically, the Locally Interpretable Layer (LIL) 
compute the relevance score for all input concepts extracted from the input sample. 
The Globally Interpretable Layer (GIL), instead, 
interpret sample predictions by means of the concepts extracted from all the training data. The GIL layer brings the advantage of studying 
how concepts from the training set influenced the classifier's decision. 
The concept scores of both layers are employed to re-predict the final class, but they are not employed for the standard prediction. 
As a baseline, RoBERTa and XLNet transformer encoders are used. GIL and LIL layers are incorporated into the transformers to make a comparison. The results reveal that incorporating GIL and LIL layers 
consistently leads to better classification across all evaluated datasets, probably due to the multitask loss employed.
To gauge the efficacy of the explanations, the authors conducted a human evaluation, where explanations provided by SelfExplain are consistently acknowledged as more understandable, trustworthy, and adequately justifying model prediction than standard XAI techniques.

\subsubsection{Prototype-based Concepts}
\label{sec:unsup_prototype}
\begin{figure}[t]
    \centering
    \includegraphics[scale=0.33]{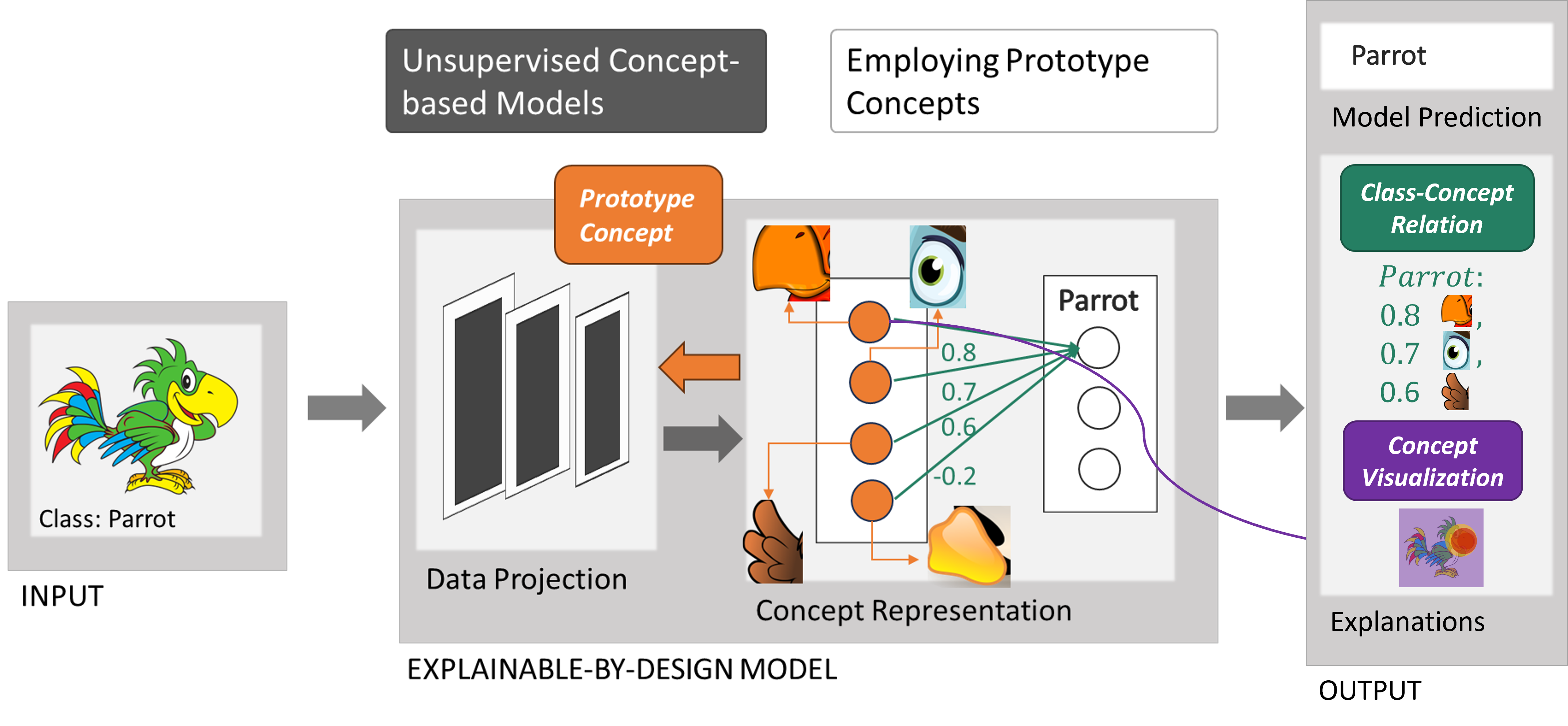}
    \caption{Unsupervised concept-based models employing prototype concepts.
    The network explicitly encodes prototype concepts representing (part of) training examples. The explanations are typically provided as class-concept relation and/or concept visualization.}
    \label{fig:cbm_unsup_proto}
\end{figure}

Within the domain of unsupervised concept-based models, another possibility is to require the network to explicitly encode \textit{prototype-based concepts}, which represent specific (parts of) train examples. As shown in Figure~\ref{fig:cbm_unsup_proto}, the prototypes are  
compared to input samples, and their similarities are employed as sufficient statistics to provide the final prediction.   
How the prototypes are extracted is different among the methods, either by means of an autoencoder (Prototype DL), or per class by means of the same convolutional filters (ProtoPNets), sharing them among the classes (ProtoPShare, ProtoPool) with a hierarchy of prototypes (HPNet), with deformable filters (Deformable ProtoPNets) or through an interactive process with human experts (ProtoPDebug). 
In most cases, a linear layer is employed on top of the prototypes to classify the final class and provide an interpretable class-concept relation. All the methods provide different systems to visualize the concepts since prototype concepts particularly require a visualization.    

\smallskip
\textbf{Prototype DL~\cite{li2018deep}} introduces the use of prototypes in concept-based explainability. 
The proposed model explains predictions based on similarity to prototypical observations within the dataset. 
Its architecture comprises an autoencoder and a prototype-based classification network.
The autoencoder aims to reduce input dimensionality and learn predictive features. 
The autoencoder's encoder facilitates comparisons within the latent space, while the decoder enables the visualization of learned prototypes. 
Given the encoded input, the prototype classification network generates a probability distribution across K classes. This network includes a prototype layer where each (prototype) unit stores a weight (prototype) vector mirroring an encoded training input.
For model prediction interpretation, we can visualize the learned prototype by feeding prototype vectors into the decoder. Then, from the learned weights of the prototype layer, we can identify which prototypes are more representative of a class. 
The experimental results show that the proposed interpretable architecture does not compromise predictive ability compared to traditional and non-interpretable architecture.

\smallskip
Prototype Parts Network \textbf{(ProtoPNet~\cite{chen2019looks})} identifies different parts of the input image that look like some learned prototypical parts.
Similarly to \textit{Li et al.}~\cite{li2018deep}, ProtoPNet incorporates a prototype layer but employs convolutional layers for the prototype extraction process.
Thus, ProtoPNet extracts prototypes based on segmented image parts rather than the whole images. Specifically, the convolutional layers extract a set of features for each image patch. ProtoPNet then computes the similarity between each prototype unit and all the image patches.
This process yields a similarity score activation map (similar to a saliency map), where each value indicates the presence of a prototypical component for each part of the image.
These similarity maps generated by individual prototype units are then condensed into a single similarity score for each prototype. 
The final prediction is based on a weighted combination of similarity scores according to the importance of a prototype for a given class. 
The resulting model provides explanations in three forms: \classconcrel{} according to the weight connecting each prototype to a class, \nodeconcass{} since each prototype represents a unit of the network, and \concvisual{} through the similarity activation map. 
ProtoPNet achieves a slightly lower accuracy (up to $3.5\%$) than non-interpretable baselines. 

\smallskip
The Hierarchical Prototype network \textbf{(HPnet~\cite{hase2019interpretable})} builds upon ProtoPNet but supports a hierarchical organization of prototypes.
The model supports hierarchical classification, i.e., a classification based on a taxonomy at multiple granularities. The approach learns a set of prototypes for each taxonomy class, from coarser ones (e.g., animal) to finer ones (e.g., cat).
HPnet architecture, after the convolution layers as in~\cite{rymarczyk2021protopshare}, includes a prototype layer for each parent node of the taxonomy. Each layer includes and classifies the prototypes of the finer classes of the corresponding parent one. The design of the prototype layer is close to that of ProtoPNet and provides for the computation of the similarity scores of the learned prototype and patches of the input image.
The approach can provide \concvisual{} via a heat map showing localized portions of the input image that strongly activate the prototype at multiple levels of the hierarchy. Hence, it enables the understanding of the prediction at multiple granularities.
Thanks to its ability to handle hierarchical classification, HPNet addresses the novel class detection problem by classifying data from previously unseen classes at the appropriate level in the taxonomy. 
The experiments show that HPnet experiences a slight drop in fine-grained accuracy with respect to a black-box model. 

\smallskip
Prototypical Part Shared network \textbf{(ProtoPShare~\cite{rymarczyk2021protopshare})} is an extension of ProtoPNet that introduces the idea of sharing prototypes between classes to reduce the number of prototypes. A reduced number of prototypes, which are, as in ProtoPnet, representations of prototypical parts, improves the interpretability of predictions.
ProtoPshare addresses two issues of ProtoPNet. First, ProtoPNet derives many prototypes since each is assigned to only one class. Second, two prototype representations of two distinct classes with similar semantics can be far apart in latent space, making the final predictions unstable.  
ProtoPShare addresses these issues by introducing prototype sharing between the classes via prototype pruning.
ProtoPShare has two phases: i) initial training and ii) prototype pruning. The initial training phase follows the one of ProtoPNet. The second phase consists of the \textit{data-dependent merge-pruning}.
Given the set of prototypes after initial training,  this step merges a fixed percentage of most similar prototypes. Specifically, the work introduces a data-dependent similarity function to identify prototypes with different semantics, even if they are apart in the latent space. The pruning process proceeds by computing the similarity of pairs of prototypes. If a pair is among a given percent of most similar pairs, one prototype of the pair is removed. 
Even with fewer prototypes, ProtoPShare outperforms ProtoPNet in terms of accuracy.

\smallskip
\textbf{ProtoPool~\cite{rymarczyk2022interpretable}}, inspired by ProtoPNet~\cite{chen2019looks}, tackles some of its limitations and those of subsequent works. One limitation it addresses is the assumption of separate prototypes for each class in ProtoPNet. Like ProtoPShare~\cite{rymarczyk2021protopshare}, ProtoPool reduces the number of prototypes by sharing them across different data classes.
However, ProtoPShare's merge-pruning step leads to increased training time. ProtoPool avoids the pruning by adopting a soft assignment of prototype represented by a distribution and automatic, fully differentiable prototype assignment, thus simplifying the training process.
Furthermore, ProtoPool introduces a focal similarity function that generates saliency maps more focused on salient features compared to using ProtoPnet similarity.
Additionally, ProtoPool introduces \concvisual{} via prototype projection. This involves replacing the abstract prototype learned by the model with the representation of the nearest training patch.
Compared to ProtoPNet and ProtoPShare, 
ProtoPool achieves the highest accuracy on one dataset using fewer prototypes and comparable but lower accuracy results on another one. A user study rated ProtoPool's visualization as more interpretable than other compared methods.

\smallskip
Also \textbf{Deformable ProtoPNet~\cite{donnelly2022deformable}} builds upon ProtoPNet~\cite{chen2019looks}. The work addresses the limitation of ProtoPNet and its subsequent works of using spatially rigid prototypes, which cannot explicitly consider geometric transformations or pose variations of objects.
Deformable ProtoPNet uses spatially flexible deformable prototypes.
Each prototype consists of adaptive prototypical parts that dynamically adjust their spatial positions based on the input image. 
The work builds upon previous work modeling object deformations and geometric transformations to deform prototypes, specifically on Deformable Convolutional Network~\cite{dai2017deformable}.
The experimental results show that Deformable ProtoPNet achieves state-of-the-art accuracy.

\smallskip
ProtoPNets~\cite{chen2019looks} may provide confounder factors as part-prototype explanations (e.g., textual metadata in image X-rays or background parts), limiting the model's predictive performance and generalizability.
\textbf{ProtoPDebug~\cite{bontempelli2023conceptlevel}} addresses this issue by proposing a human-in-the-loop concept-level debugger for ProtoPNets architectures. 
Human experts identify which part-prototypes are confounders by examining ProtoPNets' explanations. This information is provided as a feedback to the model and it is fine-tuned to align with the user supervision.
ProtoPDebug assesses all learned part-prototypes during each iteration, retrieves maximally activated training samples, and prompts user judgment on confounding. Confirmed confounders are added to the \textit{forbidden concepts} set. 
Conversely, high-quality part-prototypes are incorporated into the \textit{valid concepts} set. For the fine-tuning step, ProtoPDebug proposes two losses for the fine-tuning step. 
The \textit{forgetting loss} penalizes forbidden concepts while the \textit{remembering} loss encourages remembering valid ones. 
ProtoPDebug outperforms a predecessor debugger 
and ProtoPNet, enhancing test classification accuracy.




\subsection{Hybrid Concept-based Models}
\label{sec:sup_unsup_cbm}
\begin{figure}[t]
    \centering
    \includegraphics[scale=0.33]{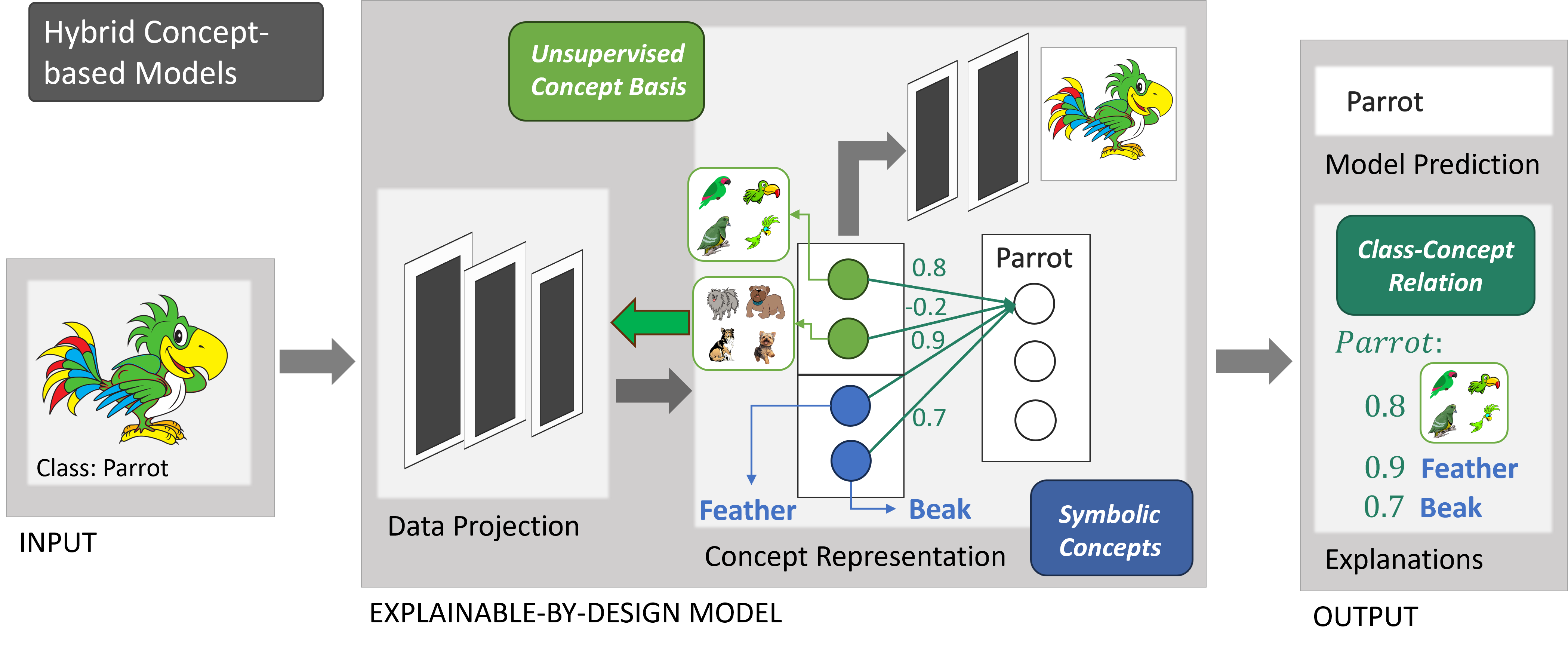}
    \caption{Hybrid concept-based model employing both supervised and unsupervised concepts. This approach normally involves employing an unsupervised learning strategy (such as reconstructing the input image) to learn an unsupervised concept basis and a few supervisions over some neurons representing symbolic concepts.}
    \label{fig:cbm_hybrid}
\end{figure}

Hybrid concept-based models propose the joint employment of both supervised and unsupervised concepts, as shown in Figure~\ref{fig:cbm_hybrid}. The strategy normally involves the employment of an unsupervised learning strategy, either reconstructing the input image with an AE (Ante-hoc, and reported in Figure\ref{fig:cbm_hybrid}) or a VAE (GlanceNet) or maximizing the mutual information (CBM-AUC), together with some supervision over some concepts. 
The goal of this integration is generally to allow the employment of concept-based models also in those scenarios where few supervised concepts are available for the task at hand.
This approach has been adopted to improve the performance of standard CBM (CBM-AUC, Ante-hoc Explainable Model) and its capability to represent concepts (GlanceNet).

\smallskip
Concept Bottleneck Model with Additional Unsupervised Concepts \textbf{(CBM-AUC~\cite{sawada2022concept})} is the first approach designed to merge CBM~\cite{koh2020concept} with SENN~\cite{alvarez2018towards}, employing supervised and unsupervised concepts at the same time. 
%
CBM-AUC addresses the issue of CBM's restricted concept efficiency. 
To address this, the authors suggest enabling the model to leverage the representation capability associated with unsupervised concepts. 
%
The first contribution is the introduction of M-SENN, a variant of SENN reducing the computational complexity and improving the scalability of SENN. 
First, it replaces the decoder of the AE architecture employed by SENN with a discriminator. This module now optimizes
the mutual information between the input and the unsupervised concepts. 
Second, M-SENN parametrizer now 
shares the intermediate network with the encoder, reducing the computational complexity. 
For integrating CBM into SENN, they add supervision on some concept basis, which then comprises unsupervised and supervised neurons.
The results show that the modifications made to join SENN with CBM are effective. 
CBM-AUC outperforms the task accuracy of both SENN and CBM. 
Also, they analyze both supervised and unsupervised concepts by visualizing their activation through saliency maps. They show that both types of concepts look at semantically meaningful areas of the input images. 

\smallskip
Similarly to CBM-AUC, also Learning \textbf{Ante-hoc Explainable Models via Concepts~\cite{sarkar2022framework}} 
integrated supervised and unsupervised concepts 
starting from SENN and CBM. 
%
Differently from CBM-AUC, however, it 
still employs SENN concept decoder to learn unsupervised concepts. 
However, the most important difference between both models is that the classifier is added on top of the base encoder and not the concepts. This entails that the final classification is not based on the concepts. 
To minimize this issue, they add a classifier to the concept encoder and minimize predictions' divergence via a fidelity loss. 
As presented, the model is designed to operate in unsupervised concept learning mode. A loss function is employed to align the learned concepts with the available annotations to allow the model to learn supervised concepts. 
Otherwise, self-supervision can also be integrated as an auxiliary task, such as predicting the rotation of the input images starting from the encoded concepts. 
The proposed framework demonstrates competitive performance across both annotated and not annotated datasets, 
improving the task accuracy of CBM and SENN, respectively. 

\smallskip
Unlike previous work, \textbf{GlanceNets~\cite{marconato2022glancenets}} aims to improve the concept representations of SENN and CBM rather than the performance. It improves the interpretability of SENN 
by employing a VAE instead of a standard AE to learn disentangled concepts. 
Due to the variational approach, learned concepts can better represent sample generative factors and respond to their alterations. 
When supervised concepts are available, they force a mapping with the learned concepts to preserve known semantics. 
They also improve the representations of CBM by identifying a solution to 
the concept leakage issue (see Section~\ref{sec:cbm_issues} for a description). They claim it is a deficiency in the out-of-distribution generalization of concept-based models. 
Therefore, GlanceNets employs an open-set recognition technique at inference time to detect instances that do not belong to the training distribution. They show this solution substantially improves the concept alignment with respect to CBM. 
Task performances are on par with CBM and SENN. 

\subsection{Generative Concept-based Models}
\label{sec:generative}
\begin{figure}[t]
    \centering
    \includegraphics[scale=0.33]{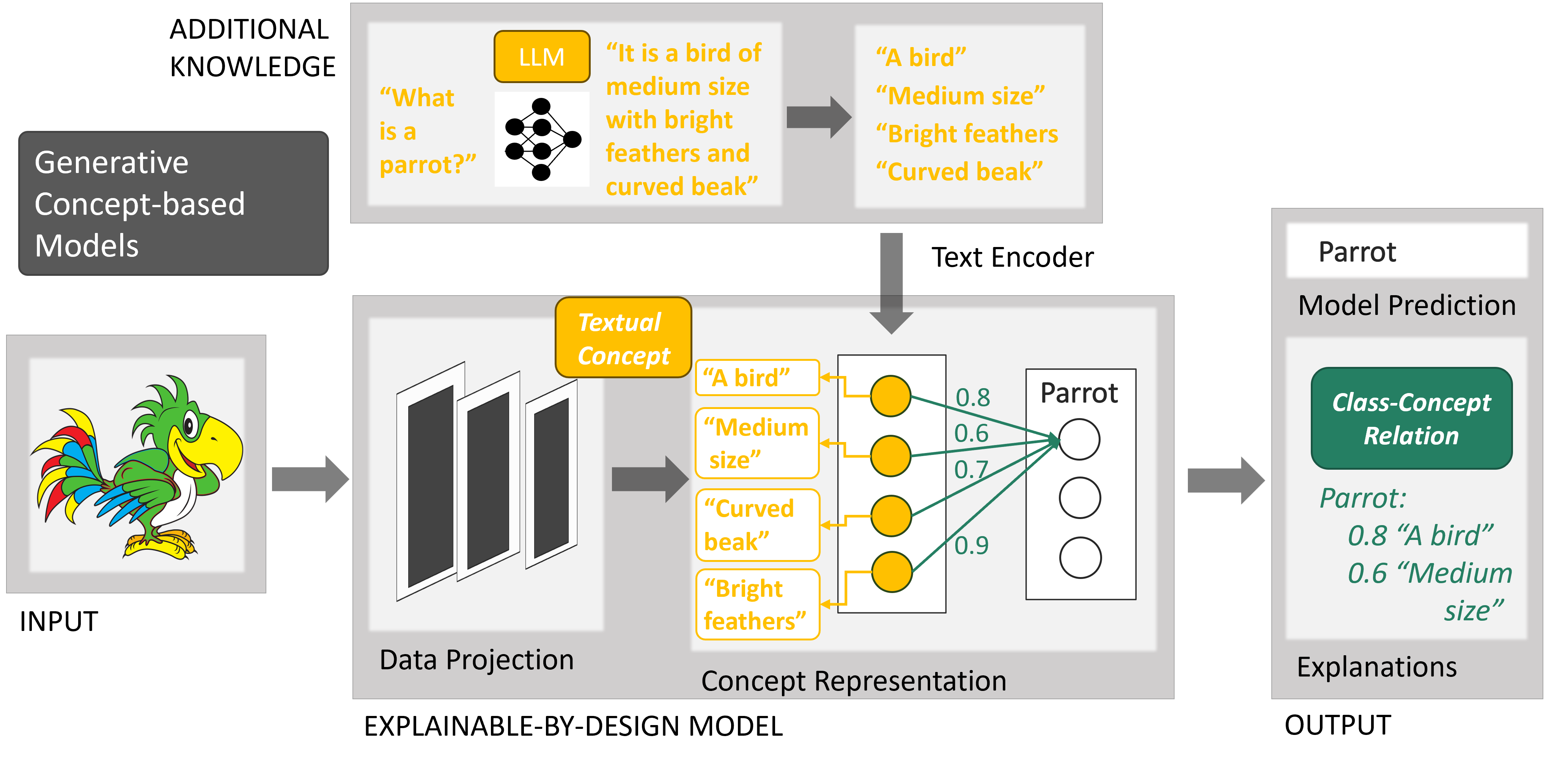}
    \caption{Generative concept-based models employ an external LLM to generate textual concepts, i.e., short descriptions of the final classes without requiring a concept annotation. The embeddings representing the textual concepts are aligned with the input latent representation. The alignment produces the concept scores, which are used to provide the final classification.}
    \label{fig:cbm_gen}
\end{figure}

Rather than employing a set of symbolic supervised concepts or extracting an unsupervised concept basis, generative concept-based models have recently shown that they can directly create concept representations. 
As shown in Figure~\ref{fig:cbm_gen}, this is generally performed by employing a Large Language Model (LLM) producing \textit{Textual Concepts}, short textual descriptions of the final class. The embeddings representing the textual concepts are aligned to the latent input representation to output concept scores. These scores are then employed to provide for the final classification, possibly with an interpretable classifier providing class-concept relations. 
At test time, these models predict both the final class and the most suitable descriptions among those that have been learned.  
In the following, we outline two methods (LaBO~\cite{yang2023language} and Label-free CBM~\cite{oikarinen2023label}) of this recent line of work. The two methods are similar in their main components since they both employ an LLM in combination with a vision-language one and provide class-concept relation explanations.

\smallskip
Language-guided Bottlenecks \textbf{(LaBO~\cite{yang2023language})} is the first model proposing this idea. It exploits the Generative Pre-trained Transformer 3 (GPT-3) language model \cite{brown2020language} to generate factual sentences to derive concepts. This approach circumvents the requirement for costly manual annotation of concepts.
In the LaBO framework, the process initiates by prompting the GPT-3 model with queries designed to generate a collection of \emph{candidate concepts} for each class. An example of a prompt is: \textit{"Describe what the <class\_name> looks like."} 
Subsequently, the approach applies a submodular optimization technique to select a subset of the generated sentences to maximize the discriminability and diversity of the concepts.
The generated textual concepts are then aligned to images using a pretrained alignment model to form a bottleneck layer. 
Specifically, LaBO uses CLIP~\cite{radford2021learning} to derive embeddings of concepts in textual form and images. 
Subsequently, a linear layer is applied to the similarity scores of concepts and images to learn a weight matrix whose values represent the importance of the concepts in the classification task.
The effectiveness of LaBo is intrinsically linked to the representation of the classes and concepts within the adopted LLM. 
LaBo retains accuracy in limited training data scenarios, but as data volume grows, its performance slightly decreases while remaining competitive compared to black-box models.
Compared to PCBM~\cite{yuksekgonul2022posthoc} as an interpretable baseline, LaBo exhibits a substantial performance improvement. 

\smallskip
Label-free Concept Bottleneck Model \textbf{(LabelFree-CBM~\cite{oikarinen2023label})} is a framework close to LaBo~\cite{yang2023language}. 
As LaBO, LF-CBM can convert a neural network into an interpretable CBM without requiring annotating concepts but by generating them using a language model. 
Its main steps also resemble LaBO ones.
It generates the concepts using GPT-3~\cite{brown2020language} and filters them to enhance the quality and reduce the concept set's size.
Then, it computes embeddings from the network backbone of input data and the concept matrix. 
The concept matrix is derived using a visual-text alignment model (CLIP-Dissect~\cite{oikarinen2022clip}). 
Each element of the matrix is the product between the CLIP encoding of the images of the training set and the concepts.
The framework then involves learning the concept weights to create the Concept Bottleneck Layer. 
Lastly, the final predictor is learned by training a fully connected sparse layer.
The authors assessed the interpretability of the model with a user study.
Regarding performance, LF-CBM demonstrates only a small loss in task accuracy compared to the original neural network backbones while it outperforms the interpretable PCBM~\cite{yuksekgonul2022posthoc}.

\subsection{Discussion}
\label{sec:cbm_issues}

In the following, we analyze critical aspects of explainable-by-design models.
We focus on three perspectives: (i) performance loss, (ii) information leakage in concept representation, and (iii) pitfalls of concept intervention.

\myparagraphbold{Performance loss}
Concept-based models explicitly represent a set of concepts within the model architecture and rely on these concepts for the final task predictions.
On the other side, the higher transparency of these architectures comes at the cost of a performance loss compared to traditional, black-box architectures.
We analyze this aspect in Table~\ref{tab:summary:bydesign1}. 
Still, we note that several approaches (that we report with \xmark) match the performance of complex black-box prediction models, with the advantage of enabling decision explanations through concepts. 

\myparagraphbold{Information leakage in supervised concept representation}
Recent works revealed that supervised concept-based models encode additional information in the concept space other than the concept information itself~\cite{havasi2022addressing, DBLP:journals/corr/abs-2105-04289, mahinpei2021promises}. 
Concept representations may include representations of the task labels when trained jointly with the downstream task. 
This phenomenon is denoted as \textit{information leakage in concept representation}, and compromises the model's interpretability and intervenability (i.e., the possibility to perform concept intervention) ~\cite{havasi2022addressing, DBLP:journals/corr/abs-2105-04289, mahinpei2021promises}.
From the interpretability perspective, we cannot directly interpret the association between the concept representations and the model decision since the outcomes may depend on other encoded information. 
From the intervenability one, interventions 
may remove the extra information encoded and can decrease the final model performance.
Mahinpei et al.~\cite{mahinpei2021promises} argue this issue is shared among concept-based models that use soft or fuzzy representations to represent concepts and experimentally demonstrate it for CBM~\cite{koh2020concept} andCW~\cite{chen2020concept}).
Mahinpei et al.~\cite{mahinpei2021promises} also show that information leakage is not avoided even when applying mitigation strategies, including (i) training concept representations and the concept-to-task sequentially (rather than jointly), (ii) adding unsupervised concept dimensions to account for additional task-relevant information, or (iii) via concept whitening~\cite{chen2020concept} which explicitly decorrelate concept representations during training~\cite{mahinpei2021promises}.
Hence, they conclude that soft representations encode concept information and data distribution.
Recent works specifically address leakage issues in concept-based models, such as~\cite{havasi2022addressing} for CBMs and GlanceNets~\cite{marconato2022glancenets}.

\myparagraphbold{Pitfalls of concept intervention}
Concept-based models enable human experts to intervene in the concept representation and modify them at test time. 
Concept intervention improves final task performance when intervening on mispredicted concepts and generating counterfactual explanations.
However, concept intervention techniques might be affected by major pitfalls. 
Shin et al. identify two major issues~\cite{shin2023closer} on CBMs~\cite{koh2020concept}.
First, intervention could increase the task error, compromising the reliability of intervention techniques. 
This decrease in performance could happen when users nullify void concepts by setting unsure concepts to zero. 
The authors justify this phenomenon, noting that the absence of a concept in a specific image does not entail the concept is not associated with its class. 
We argue this is also linked to the soft representation of concepts and leakage issues.  
The second issue arises from the practice of using majority voting to associate concept values to classes to improve task performance (e.g., in \cite{koh2020concept, zarlenga2022concept, havasi2022addressing}). 
Majority voting affects a fair representation of minority samples and makes task outcomes biased toward the majority. 
The authors argue that intervention can exacerbate this unfair treatment since it would increase minority performance only when changing the concept value to the majority one.


\section{Evaluation} 
\label{sec:evaluation}
This section overviews the quantitative and qualitative metrics, the datasets, and the resources mainly used in the literature to assess the performance of C-XAI methods. 
In Section~\ref{sec:metrics}, we delve into the quantitative metrics employed to evaluate the methodologies. This includes measures of the effect of concepts on task predictions and performance, and the quality of concepts. Conversely, Section~\ref{sec:human_eval} elaborates on the qualitative evaluations which leverage human assessments. Section~\ref{sec:datasets} provides an overview of the datasets utilized to assess the effectiveness of C-XAI methods. Finally, in Section~\ref{sec:resources}, we provide details about the resources accessible for each method.

\subsection{Quantitative evaluation} \label{sec:metrics}
We categorize the quantitative metrics used and proposed in the reviewed papers into two categories according to their final purpose and what they measure.
The metrics of the first category assess how well concepts contribute to class predictions and task performance. The second category focuses on the intrinsic quality of concepts. For a more detailed explanation of each single metric, please refer to Appendix\ref{app:metrics}.

\myparagraphbold{Concept effect on class prediction and task performance}
These metrics assess how concepts contribute to (i) the final class prediction or (ii) the task performance.

\textit{(i) Concept effect on class prediction.}
This set of metrics assesses the effect of the concepts on the final class prediction. Concept-based methods often naturally provide the concept relevance (importance) by means of the concept weight connecting the concept to the final classes. Hence, these metrics have been mostly employed for post-hoc explanation methods providing \classconcrel\ to study the impact of the identified concepts on task prediction. 
Among these metrics we find the T-CAV score \cite{kim2018interpretability, ghorbani2019towards}, CaCE~\cite{goyal2019explaining,wu2023causal},  ConceptSHAP~\cite{yeh2020completeness}, Smallest Sufficient Concepts (SSC)~\cite{ghorbani2019towards}, Smallest Destroying Concepts (SDC)~\cite{ghorbani2019towards, vielhaben2023multi}, STCE Concept Importance~\cite{ji2023spatial}, T-CAR score~\cite{crabbe2022concept}, Sobol score~\cite{fel2023craft}, Concept Weight~\cite{yuksekgonul2022posthoc, yang2023language}, Concept Relevance~\cite{alvarez2018towards,rigotti2022attentionbased, barbiero2023interpretable}.

\textit{(ii) Concept effect on task performance.}
 These metrics assess how well concepts contribute to the predictive capacity of the model. 
 They are employed in post-hoc methods, to assess the quality of the extracted concept for predicting the final class; at the same time, they are also studied in concept-based models, to assess how well concepts contribute to the predictive capacity within the given model.
 Among these metrics we find the Completeness Score~\cite{yeh2020completeness}, Fidelity~\cite{zhang2021invertible, sarkar2022framework, fel2023craft}, Concept Efficiency~\cite{zarlenga2022concept}, Conciseness~\cite{vielhaben2023multi}.

\myparagraphbold{Quality of concepts}
This second category focuses on the quality of concepts, studying (i) their inherent properties, (ii) how they relate to internal network representations, and (iii) their impact on prediction errors. In the following, we review these metrics with respect to these three targets of focus.

\textit{(i) Concepts properties.}
This group of metrics centres on the intrinsic characteristics of concepts themselves
They evaluate the retained information, distinctiveness, completeness, and purity of concepts, providing insights into the information content and coverage of concepts. Among these metrics we find the Mutual Information~\cite{zarlenga2022concept}, Concept Distinctiveness~\cite{wang2023learning}, 
Concept Purity~\cite{wang2023learning}. 
  

\textit{(ii) Relation of Concepts with internal network representation.}
This group of metrics quantifies the relationship between a node (or filter) of the model and the concept it represents. These metrics are adopted for methods that provide explanations as \nodeconcass{}. They assess the spatial overlap (IoU) and location stability of concepts within activation maps. Among these metrics we find Intersection over Union \cite{bau2017network, fong2018net2vec, xuanyuan2023global} (and adopted~\cite{zhang2018interpretable} with the name of \textit{Part Interpretability}), Location stability~\cite{zhang2018interpretable}.


\textit{(iii) Concept prediction error.}
This group of metrics measures the alignment between learned concepts and concept ground truth. Hence, they are applicable when concept labels are available. These metrics evaluate and quantify mispredictions and the purity of concepts. Among these metrics we find the Concept Error~\cite{koh2020concept} (and used in~\cite{sarkar2022framework} with the name Explanation Error), AUC score~\cite{chen2020concept}, Misprediction-overlap metric (MPO)~\cite{kazhdan2020now}.


\subsection{Qualitative evaluation}\label{sec:human_eval}
As common in XAI assessment~\cite{bodria2023benchmarking}, several C-XAI works perform human evaluations to assess their method through human feedback.
We report which methods perform human evaluations in the ``Human evaluation'' column of Tables \ref{tab:summary:posthoc2} (post-hoc methods) and \ref{tab:summary:bydesign2} (by-design approaches). Half of the post-hoc approaches conduct qualitative assessments, while only a quarter of explainable-by-design ones perform them.
In the following, we overview the assessment based on the nature of the aspect under study: understandability, coherence, and utility of concept explanations.

\myparagraphbold{Understandability of explanations} 
When evaluating explanations, the predominant focus involves assessing how understandable and reasonable the explanations are to humans.
Various measures and evaluation settings have been introduced to capture specific nuances within this context. One relevant property of explanation is their \textit{plausibility}, denoting the extent to which explanations align with human reasoning.
One direct setting for evaluating the explanation plausibility consists of asking users to determine if the explanation is justifiable and understandable based on their expectation~\cite{rajagopal2021selfexplain, bau2017network}.  
To evaluate \textit{human understanding} of concepts, another evaluation approach requires users to label the identified concepts by selecting some phrases from a predefined vocabulary that best describes the concept~\cite{wang2023learning}. This method provides a structured measure for summarizing users' comprehension of identified concepts. Other approaches as~\cite{zhang2021invertible} measure the interpretability of a concept by asking users to provide free-text labels for concepts and identify, among a set, the candidate explanations matching the concept and measure their accuracy as the percentage of correctly identified concept explanations. Close to this setting,~\cite{zhou2015object} focus on quantifying the semantics learned by each network unit and evaluating the accuracy of human-assigned text annotations to concepts.
The \textit{trustability} of explanations is another critical aspect, where users provide feedback on their level of trust in the explanation~\cite{rajagopal2021selfexplain}.
The \textit{reasonability} of explanations evaluates the effectiveness of conveying meaningful information through explanations.
A way to evaluate it is by investigating what information saliency maps can communicate to humans~\cite{kim2018interpretability}. 
Another setting consists of providing different explanations for the same prediction and asking human raters to indicate which is more reasonable, as adopted by IBD~\cite{zhou2018interpretable}.
\textit{Factuality} assesses the accuracy of concepts by having annotators judge their alignment with ground truth images~\cite{yang2023language}.

\myparagraphbold{Coherence of concepts} Another aspect to evaluate is the coherence of concepts as perceived by humans.  This involves scrutinizing and evaluating the characteristics of the discovered concepts. 
An adopted evaluation method for measuring concept coherence is the intruder detection experiment. In this setup, for example, users are asked to identify an image among a set that is conceptually different from the rest~\cite{ghorbani2019towards, fel2023craft}. Another setting involves examining the nearest neighbor images associated with each discovered concept vector and selecting the most prevalent and cohesive concept~\cite{yeh2020completeness}.
The work in~\cite{yang2023language} defines the \textit{groundability} to measure the consistency of the vision-language model adopted in their framework to ground concepts to images in a manner that aligns with human interpretations.

\myparagraphbold{Utility of explanations }
The \textit{utility} of an explanation refers to its practical usefulness and effectiveness in providing insights or guidance to users, typically in the context of the specific task.
It is often measured by its ability to assist users in predicting, performing a task, or making informed decisions in a given context.
For example, in~\cite{jain2022extending}, users evaluate the utility of explanations in improving classifier generalization through feature engineering. They identify problematic features using global explanations and evaluate the ease of use of the explanations for this task. In addition, users select the most general classifier based solely on the global explanations of various methods.
Other user study settings, such as in~\cite{fel2023craft}, measure how well the explanations assist users in inferring the insights guiding classifications. The utility is evaluated by measuring users' accuracy in predicting model decisions on novel images given the explanations.

\begin{table}[t]
\centering
\caption{Datasets employed by the reviewed methods in terms of type of data, size, task label and concepts. We report only the dataset in which the concepts are explicitly defined.  All datasets have been defined for classification tasks. We report with $^*$, the datasets suitable for regression too.}
\label{tab:datasets}
\resizebox{\textwidth}{!}{
\begin{tabular}{ll|l|l|l|l|l}

\multirow{50}{*}{\rotatebox[origin=c]{90}{\textbf{Type of Data}}} & &\textbf{Dataset} & \textbf{Used in}& \textbf{Size} & \textbf{Label} &\textbf{Concepts} \\ \hline

& \multirow{25}{*}{\rotatebox[origin=c]{90}{IMG}} & MNIST~\cite{lecun1998mnist} & ~\cite{crabbe2022concept}~\cite{alvarez2018towards} ~\cite{wang2023learning}~\cite{li2018deep} &  60000 & Digits & Geometrical attributes 
\\ 
& & MNIST even/odd~\cite{barbiero2022entropy} & ~\cite{rigotti2022attentionbased} & 60000 Even/odd & Digit'parity & Digit value \\

& & Fashion MNIST~\cite{xiao2017fashion} & ~\cite{li2018deep} & 60000
Clothes & Details of clothes \\
& & colored-MNIST~\cite{kim2019learning} &~\cite{goyal2019explaining} & 60000 & Digits & Digit's colours  \\

& & \makecell[l]{Caltech-UCSD\\ Birds-200 (CUB)~\cite{wah2011caltech}} & \makecell[tl]{~\cite{crabbe2022concept} ~\cite{zhang2021invertible} 
~\cite{leemann2023are}
~\cite{koh2020concept}\\
~\cite{kim2023probabilistic}
~\cite{kazhdan2020now}
~\cite{zhang2018interpretable}
~\cite{wang2023learning}\\
~\cite{chen2019looks}
~\cite{rymarczyk2021protopshare}
~\cite{rymarczyk2022interpretable}
~\cite{donnelly2022deformable}\\
~\cite{sawada2022concept}
~\cite{sarkar2022framework}
~\cite{yang2023language}
~\cite{oikarinen2023label}\\
~\cite{zarlenga2022concept} }& 11788 & Bird species & \makecell[l]{Visual attributes\\of the birds }\\

& & BRODEN ~\cite{bau2017network}& ~\cite{bau2017network}
~\cite{zhou2018interpretable}
~\cite{fong2018net2vec} & 63305 & Concepts & 1197 visual concepts. \\

& & CelebA~\cite{Liu_2015_ICCV} & 
~\cite{goyal2019explaining}   ~\cite{marconato2022glancenets}
~\cite{zarlenga2022concept}   ~\cite{barbiero2023interpretable} & 202599 & Person's visual attributes & \makecell[l]{Non-overlapping \\person visual attributes} \\

& & \makecell[l]{OsteoArthritis\\Initiative (OAI)$^*$~\cite{oai_website}} & ~\cite{koh2020concept} & 36369 & Osteoarthritis levels & \makecell[l]{Knee conditions\\relative to the osteoarthritis}\\

& & Places365~\cite{zhou2017places} &
~\cite{zhou2015object}
~\cite{chen2020concept}
~\cite{oikarinen2023label} & 10million & Scenes & Objects landmarks\\

& & \makecell[tl]{Animals with Attribute\\(AwA and AwA2)~\cite{awa2_website}} & 
~\cite{yeh2020completeness}
~\cite{kim2023probabilistic}
~\cite{sarkar2022framework} & 37322 & Animals & 85 numeric attribute \\

& & HAM10000~\cite{tschandl2018ham10000} & ~\cite{yuksekgonul2022posthoc}
~\cite{yang2023language} & 10000 & Malignant/Benignant & \makecell[l]{Characteristics of\\malignant skin lesions}\\

& & SIIM-ISIC~\cite{isic_challenge_2020} & ~\cite{yuksekgonul2022posthoc} & 33126 & Malignant/Benignant & \makecell[tl]{Characteristics of\\malignant melanoma}\\

& & MPI3D~\cite{gondal2019transfer} & ~\cite{marconato2022glancenets} & >1million & Objects  & \makecell[l]{Object attributes and \\robot's sensory measurements}\\

\cline{2-7}
& \multirow{6}{*}{\rotatebox[origin=c]{90}{TXT}} & CEBaB~\cite{abraham2022cebab} & ~\cite{wu2023causal} &>15000& Restaurant reviews' sentiment & Aspect-level
sentiment \\

& & IMBD~\cite{imdb_datasets} & ~\cite{yeh2020completeness} & 50000 & Movie reviews' sentiment (pos/neg) & Extracted\\

& & SST-2~\cite{socher2013recursive} & ~\cite{rajagopal2021selfexplain} & >70000 & Movie reviews' sentiment (pos/neg) & Extracted\\

& & SST-5~\cite{socher2013recursive} & ~\cite{rajagopal2021selfexplain} & >11000 & Movie reviews' sentiment (5 classes)  & Extracted\\

& &TREC-6/50~\cite{li2002learning} & ~\cite{rajagopal2021selfexplain} & >5000 & Question types (6 classes/50 classes) & Extracted\\

& & SUBJ~\cite{pang2005seeing} & ~\cite{rajagopal2021selfexplain} & 9000 &  Subjective/objective. &Extracted\\

\cline{2-7}
& \multirow{4}{*}{\rotatebox[origin=c]{90}{VID}}
& Kinetics-770~\cite{carreira2019short} & ~\cite{ji2023spatial}  & 65000 & Human actions (700 classes) & Extracted \\

& & KTH Action~\cite{schuldt2004recognizing} & ~\cite{ji2023spatial}  & >2000 & Human actions (6 classes)  & Extracted\\

& & BDD-OIA~\cite{xu2020explainable} & ~\cite{sawada2022concept}  &>22000 & Human actions (4 classes) & \makecell[l]{21 concepts\\ (e.g. Traffic light is green, etc.) }\\

\cline{2-7}
& \multirow{3}{*}{\rotatebox[origin=c]{90}{TAB}}
& COMPAS~\cite{compas_analysis} &~\cite{alvarez2018towards} & >60000 & Recidivism (yes/no) & Table attributes \\

& & V-Dem~\cite{pemstein2018v} &~\cite{barbiero2022entropy, ciravegna2023logic} & 202 &State Democracy level  & Table attributes\\

& & MIMIC-II~\cite{saeed2011multiparameter} & ~\cite{barbiero2022entropy, ciravegna2023logic} & >40000 & Patient survival (dead/alive) & Table attributes \\

\cline{2-7}
& \multirow{1}{*}{\rotatebox[origin=c]{90}{TS}} 
& \makecell[l]{MIT-BIH \\ Electrocardiogram \\ (ECG)~\cite{moody2001impact}} & ~\cite{crabbe2022concept} & 47 (187 timesteps) & Heartbeat (normal/abnormal) & \makecell[l]{Heart issues} 
\\

\cline{2-7}
& \multirow{6}{*}{\rotatebox[origin=c]{90}{GRAPH}}
& MUTAG~\cite{debnath1991structure} & ~\cite{xuanyuan2023global} & 188 & Node labels (7 classes) & Extracted \\

& & Reddit-binary~\cite{kipf2016semi} & ~\cite{xuanyuan2023global}  & 2000 & \makecell[l]{Type of graphs \\(question/answer-based community or \\ a discussion-based community)} & Extracted \\

& & PROTEINS~\cite{borgwardt2005protein} & ~\cite{xuanyuan2023global} & >1000 & Type of protein (enzymes/not enzymes) & Extracted \\

& & IMBD-Binary~\cite{yanardag2015deep} & ~\cite{xuanyuan2023global} & 1000 & Movies & Extracted \\

\end{tabular}
}
\end{table}

\subsection{Datasets}\label{sec:datasets}
We now analyze the most interesting datasets employed in the literature. Among the datasets experimented by the method reviewed in this survey, we report in Table~\ref{tab:datasets} only those where the authors either explicitly describe the concept annotations or the procedure followed to extract them (Extracted).
We classified them based on the data type: images (IMG), text (TXT), videos (VID), tabular (TAB), time series (TS), and graph data (Graph).

On top of Table~\ref{tab:datasets}, we report the image datasets.
We observe that the CUB dataset is the predominant choice in the C-XAI literature. This dataset is very suitable for such methodologies due to its extensive annotation of concepts, representing the visual attributes inherent to different bird species. 
We also note that the BRODEN dataset~\cite{bau2017network} was introduced explicitly for concept-based XAI. This dataset encompasses over 63,000 images and 1,197 concepts distributed across six categories (i.e., textures, colors, materials, parts, objects, and scenes).
Finally, OAI is the only dataset that is suitable for regression tasks and not only for classification (the level of osteoarthritis may be both considered as a class or as a continuous level).
%
In the second part, we find the text datasets. These datasets predominantly revolve around binary or multi-class sentiment classifications. In most datasets, the concepts are not already present but are extracted from the input texts using various techniques. For example, in~\cite{rajagopal2021selfexplain}, concepts are represented by non-terminal phrases of the parsing tree.
%
Within the third part of the table reporting videos dataset, BDD-OIA is the only dataset with pre-established concepts, while in the other two cases, concepts are extracted through specific processes.
%
In tabular data sets, data attributes are used concepts. Finally, in graph datasets, concepts are always extracted by the method.

\subsection{Resources}\label{sec:resources}
We now report information and insights regarding the resources associated with the reviewed methods. 
Specifically, when we refer to \textit{resources}, we are addressing whether the authors of the respective papers have made the code, models utilized in their research publicly accessible or have released a new dataset. 
Among all methods discussed, nearly all but four have made their code publicly available on GitHub, two post-hoc methods (CaCE, Object Detection) and two concept-based models (SENN, CBM-AUC).
Overall, there is a common trend of releasing code on public platforms (e.g., GitHub), which benefits the community by providing a starting point for testing and development.  
Concerning data release, four methods released novel datasets, equally distributed between post-hoc methods (ND, DMA \& IMA) and concept-based models (CME, Glance-Nets).
The release of new datasets is infrequent due to the resource-intensive nature of creating concept annotated datasets. It may require $C\times N$ extra annotations where $C$ represents the number of concepts and $N$ the number of samples. In other cases, a class-level annotation of the concept is preferred to decrease this effort. It consists in employing the same concept annotation for all the samples belonging to the same class, therefore reducing the number of annotations to $N\times Y$, where $Y$ represents the number of final classes.
For additional information on the works with shared code, models, or datasets released, please consult Appendix \ref{app:resources}.


\section{Applications}
\label{sec:application}
The recent emergence of C-XAI methods has 
led to some interesting applications. We report first novel algorithms based on concepts in the field within the AI area (outside explainability), followed by real-world applications.

\myparagraphbold{Support to AI algorithm's development}
The first application of concept-based methods within AI came out in the image generation field. In VAEL~\cite{misino2022vael}, the authors proposed to condition the generation of images by means of a neuro-symbolic program working over concepts. 
For example, given two pairs of digits from MNIST, they train the concepts to represent the digits and the logic program to predict the addition of the digits. When decoding, they require instead that the generated images correspond to samples solving the given query (e.g., digits whose sum is 13). Interestingly, they show that their method also generalizes to queries unseen in training (e.g., they generate the samples whose difference or factorization is provided). 

Other concept-based algorithms appeared in the out-of-distribution sample detection field. In particular, the author of~\cite{madeira2023zebra} proposed a concept-based approach to characterize outliers. They propose first mapping the sample into an interpretable feature space comprising human-defined concepts like textures, colors, and geometry. In this space, they analyze the rarity of the features to detect outliers and explain the rarest concepts identified in the sample.
The author of~\cite{choi2023concept}, instead, employs an unsupervised approach to extract concepts that can sufficiently characterize the decisions of an out-of-distribution detector. Furthermore, they require extracted concepts to separate in-distribution and out-of-distribution samples. Interestingly, they show that extracting concepts to explain the classifier is insufficient to explain the out-of-distribution detector's prediction.

\myparagraphbold{Real-world applications}
Outside the AI world, the medical sector is where we find most applications of C-XAI already in the reviewed papers (CAR, CBM, PCBM, ProtoPDebug, LaBO). Another example is~\cite{lucieri2020interpretability}, where the authors employ a post-hoc concept-based method in the Computer-Aided Diagnosis (CAD) systems in skin cancer diagnosis. 
The authors want to elucidate the basic principles that guide the model's predictions, discerning whether the model operates on similar concepts to those used by dermatologists to describe and diagnose the disease. Specifically, using TCAV~\cite{kim2018interpretability} method, they show a strong correlation between DNN's learned representation of concepts and those routinely used by dermatologists. Hence, listing the concepts that positively influenced the final outcome may be sufficient for trusting the predictions of a CAD system. 

In the field of education, we identify RIPPLE ~\cite{asadi2023ripple} as a first application. This work introduces a pipeline to make predictions from raw time series data about students' success that is also interpretable. Specifically, they use a combination of a graph-based neural network approach for classifying raw time series of student interactions and an adaptation of concept activation vectors TCAV~\cite{kim2018interpretability} for interpreting NN internal state in terms of human-interpretable concepts. In this case, concepts are represented by six learning dimensions from educational scenarios defined a-priori.

\section{Promising Directions}
\label{sec:future_work}
The analysis of the C-XAI field literature allows highlighting  several promising directions existing for both researchers and practitioners. Addressing these frontiers is important not only for enhancing the interpretability of artificial intelligence systems but also for fostering their responsible and effective deployment across diverse domains. Herein, we outline key directions that we believe merit focused explorations, encompassing both methodological advancements and improvements from an evaluation and application point of view.

From a methodological point of view, we believe that generative approaches could also be explored for post-hoc concept-based methods. On one side, Generative models could be harnessed to create concepts akin to their role in concept-based models. On the other, LLMs could also be employed to enhance supervised post-hoc methods in assessing model causality when prompted with specific concepts.
Still considering the post-hoc methods, the analysis of Table~\ref{tab:summary:posthoc1} reveals a gap in exploring node-concept association explanations for unsupervised methods. Even though it may be difficult to associate nodes to extract concepts, it represents a valuable avenue for exploration.
Concerning concept-based models, extending concept representations to accommodate multi-modal data, such as text, images, and audio, could greatly enhance the versatility of C-XAI across different domains.
Also, the novel generative approach seems open to several extensions. Moreover, the new generative approach seems to be open to different types of fusion following the same approach: a concept generator that creates representations (not necessarily textual) that can be used in an external (not necessarily visual) model. 


On the evaluation front, the field currently lacks benchmark datasets and standardized metrics, hindering systematic comparisons. Developing these resources would facilitate fair assessments of different methods. 

From an application point of view, it would be important to develop scalable concept-based methods and models that can handle and be integrated into large-scale models and applied to large-scale datasets. 
Finally, properly integrating concept-based models is crucial to creating accountable AI-powered decision-making systems, particularly in critical domains like healthcare, finance, and autonomous systems.

\section{Conclusion} 
\label{sec:conclusions}

In conclusion, our comprehensive review of Concept-based explainable Artificial Intelligence highlights the growing interest in concept-based explanations in addressing the demand for transparent and interpretable AI models. The ongoing debate regarding the limitations of raw feature-based explanations has led to a substantial body of work on concept-based explanations. 
By defining the various types of concepts and concept-based explanations and introducing a systematic categorization framework, we hope to have facilitated a more structured understanding of the C-XAI landscape. Our analysis of C-XAI papers, which we reviewed and categorized with the proposed taxonomy into nine categories, may provide valuable insights into the strengths and challenges of different categories. 
Furthermore, by highlighting key resources within the C-XAI literature, such as datasets and metrics, we aspire to catalyse future research and facilitate comparisons between different methodologies. 


\bibliographystyle{acm}
\bibliography{main}

\appendix


\section{Glossary of terms}
\label{app:glossary}
In the following, we provide a glossary of the most important term in the field of Concept-based XAI. We hope this glossary will serve readers as a unique reference source, allowing future literature to use the different terms consistently. For the sake of readability, we will always reintroduce these terms upon their first mention. In Table~\ref{tab:acronyms}, instead, we provide a list of the acronyms and abbreviations.

\begin{itemize}[leftmargin=3.2cm]
\item[Explainability] The capacity of a method to provide intelligible explanations for model predictions or decisions. It should unveil the inner workings and factors that contribute to the model's outcomes in a manner that is understandable to humans.
\item[Transparency] The characteristics of a model whose inner processes, mechanisms, and logic are clear and comprehensible to users. A transparent model should reveal how it reaches specific predictions, enhancing user trust and aiding in identifying potential biases or errors.
\item[Interpretability] It is the degree to which the outcomes and decision-making of a machine learning model can be grasped and reasoned about by humans. An interpretable model reveals the relationships between input data and predictions, facilitating a deeper understanding of its functioning.
\item[Local Expl.] It focuses on interpreting the behavior of a model for a specific individual prediction. It provides insights into why the model made a particular decision for a given input instance.
\item[Node/Filter Expl.] It involves detailing the behavior of individual nodes or neurons within a neural network. It helps decipher the specific features or relationships that targeted nodes consider during predictions.
\item[Global Expl.] It provides insights into a model's behavior across the entire dataset. It identifies patterns, trends, and overall tendencies to understand better how the model functions holistically.
\item[Post-hoc Expl.] Explanations are generated after the model has made a prediction. They do not interfere with the model standard working. 
\item[Model Agnostic] Technique that is not dependent on the specific algorithm or model architecture used. Model-agnostic approaches can be applied to various machine learning models by only analyzing their input and outputs.
\item[Counterfactual Expl.] Explanations showing how altered input data could change model predictions. They create modified scenarios to show why a model made a specific prediction and how input changes would result in different outcomes.
\item[Feature Importance] It measures the contribution of each input feature to a model's predictions. It helps identify which features have the most significant influence on the model's outcomes (e.g., pixel images).
\item[Saliency Map] A saliency map is a visual representation highlighting the important regions or features of an input that significantly affect a model or a node's behavior. 
\item[Concept] High-level, human-understandable abstraction that captures key features or patterns within data (e.g., object parts, colors, textures, etc.). Concepts are used to simplify complex information, enabling easier interpretation and communication of AI model behavior.
\item[Concept-based Expl.] Type of explanations aiming to elucidate model predictions by decomposing them into human-understandable concepts. Rather than providing explanations regarding feature importance, these methods provide explanations in terms of high-level entity attributes (i.e., concepts). These explanations aim to bridge the gap between complex model decisions and intuitive human reasoning.
\item[Concept-based Model] Type of machine learning model designed to make predictions while employing high-level concepts or abstractions rather than working directly with raw data. These models enhance interpretability by relying on human-understandable concepts.
\item[Concept Bottleneck] Design principle of certain concept-based models. It involves incorporating an intermediate layer within the model's architecture to represent a given set of concepts. This layer constrains the flow of information by forcing the model to represent its predictions using these predefined concepts, possibly decreasing the model classification performance.
\item[Concept Score]  Numerical value that quantifies the relevance or presence of a specific concept within the context of a machine learning model's decision or prediction. 
\item[Concept Embedding] Representation of concepts into a numerical form. Concept embeddings map complex human-understandable concepts into numerical representations, providing a model of more information than what can be described by a single score.
\item[Logic Expl.] An explanation outlining the step-by-step logical reasoning process leading to a particular decision or prediction. It presents a sequence of logical conditions (i.e., a rule) used by the model to reach its conclusion, e.g., $paws(x) \land tail(x) \land muzzle(x) \rightarrow dog(x)$. 
\item[Probing Method] Technique to assess how the latent representations captured by a given model can discriminate a set of concepts. This method involves training an auxiliary concept-specific model on top of the neural network's latent representation. Probing methods enable researchers to uncover the underlying concepts learned by deep models. 
\item[Concept Intervention] Technique to assess and modify a machine learning prediction based on modification of the predicted concepts. They enable domain experts to test hypothetical scenarios to improve the model's prediction and provide concept-based counterfactual explanations.
\end{itemize}
\begin{table}[t]
    \centering
    \begin{tabular}{l|l}
    \textbf{Acronym}         &   \textbf{Full name} \\
    \hline
        ACE \cite{ghorbani2019towards}         &	Automatic Concept-based Explanations \\
        AE          &   Auto Encoder \\
        AI          &	Artificial Intelligence \\
        CAM \cite{zhou2016learning}         &	Class Activation Map \\
        CAV \cite{kim2018interpretability}         &	Concept Activation Vector \\
        CAR \cite{crabbe2022concept} & Concept Activation Region \\
        CBM \cite{koh2020concept}         &	Concept Bottleneck Models \\
        CEM \cite{zarlenga2022concept}         &	Concept Embedding Models \\
        CNN         &	Convolution Neural Network \\
        CT \cite{rigotti2022attentionbased} & Concept Transformer \\
        CW \cite{chen2020concept} &   Concept Whitening \\
        C-XAI        &   Concept-based XAI \\
        DCR \cite{barbiero2023interpretable}  &  Deep Concept Reasoning \\
        DeconvNet \cite{zeiler2014visualizing}   &	Deconvolutional Network \\
        DL          &	Deep Learning \\
        DNN         &	Deep Neural Network \\
        GDPR        &	General Data Protection Regulation \\
        Grad-CAM \cite{selvaraju2017grad}    &	Gradient-weighted Class Activation Mapping \\
        GNN         & Graph Neural Network \\
        IBD \cite{zhou2018interpretable}         & Interpretable Basis Decomposition \\
        ICE \cite{zhang2021invertible}  & Invertible Concept-based Explanation \\
        LaBO \cite{yang2023language} & Language in a bottle
        \\
        LEN \cite{ciravegna2023logic} & Logic Explained Networks \\
        LIME \cite{ribeiro2016should}        &	Local Interpretable Model-agnostic Explanation \\
        LLM         &   Large Language Model \\
        ML          &	Machine Learning \\
        MLP         &	Multi-Layer Perceptron \\
        ND \cite{bau2017network}          &   Network Dissection \\
        Net2Vec \cite{fong2018net2vec}    &   Network to Vector \\
        PCA         &	Principal Component Analysis \\
        PCBM \cite{yuksekgonul2022posthoc} & Post-hoc Concept Bottleneck Models \\
        ProtoPNets \cite{chen2019looks} & Prototype Parts Networks \\ 
        SENN \cite{alvarez2018towards}        &	Self-Explaining Neural Network \\
        SHAP \cite{lundberg2017unified}        &	SHAPley additive explanation \\
        TCAV \cite{kim2018interpretability}        &	Testing with CAV \\
        VAE         &   Variational Auto Encoder \\
        XAI         &	EXplainable AI \\
    \hline
    \end{tabular}
    \caption{Table of acronyms and abbreviations.}
    \label{tab:acronyms}
\end{table}

\section{Formal Concept Analysis}
\label{sec:formal_concept}
We now formally define the notion of concepts, using as reference the category theory literature~\cite{goguen2005concept, ganter2012formal, zhang2006approximable}. This area has been examined the topic from various perspectives, providing a robust base for our definition.
A concept is a set of attributes that agrees on the intention of its extension~\cite{zhang2006approximable}, where the extension of a concept consists of all objects belonging to the concept and the intention of a concept is the set of attributes that are common to all these objects. 

More precisely, consider a context defined as the triple $(O, A, I)$, where we indicate with $O$ a set of objects, with $A$ a set of attributes, and with $I$ the satisfaction relation (also called “incidence”) which is a subset of $A \times O$. We also define a subset of objects $\hat{O} \subset O$ and a subset of attributes $\hat{A} \subset A$, a derivation operator $\hat{A}' = \{o \in O | (o,a) \in I, \forall a \in \hat{A}\} $ and dually $\hat{O}' = \{a \in A | (o,a) \in I, \forall o \in \hat{O}\} $. We define $(\hat{A}, \hat{O})$ as a Formal Concept if and only if $\hat{B}'=\hat{A}$ and $\hat{A}' = \hat{B}$, or, also, if $\hat{A}'' = \hat{A}$ and $\hat{O}''=\hat{O}$. In other words, a concept is a set of attributes that are shared among a set of objects and for which no other attribute exists that is shared among all the objects, even though other attributes can characterize some objects.
According to~\cite{goguen2005concept}, this is the broadest possible concept definition. 
Indeed, it allows us to consider all the previously identified typologies of concepts according to what we consider as the set of attributes $A$.  

\section{Metrics Description}
\label{app:metrics}
We now deep the analysis provided in Section\ref{sec:metrics} with a short description of each metric within each of the categories outlined.

\subsection{Concept effect on class prediction and task performance}

\myparagraphbold{(i) Concept effect on class prediction}

\begin{itemize}
    \item T-CAV score: it is the fraction of the class’s inputs with a positive conceptual sensitivity~\cite{kim2018interpretability, ghorbani2019towards}, where the conceptual sensitivity is how much each concept influences the prediction of a given class. So, it measures how many samples have concepts whose effects lie positively on the final prediction. 
    \item CaCE: it measures the causal effect of the presence of concepts on the final class prediction~\cite{goyal2019explaining,wu2023causal}. 
    \item ConceptSHAP: it assesses the effect of a concept on the final completeness score~\cite{yeh2020completeness}. The completeness is defined as the amount to which concept scores are sufficient for predicting model outcomes. 
    \item Smallest Sufficient Concepts (SSC):
    it computes the class accuracy when employing only a subset of the concepts, looking for the smallest set of concepts~\cite{ghorbani2019towards}. 
    \item Smallest Destroying Concepts (SDC): it looks for the smallest set of concepts deleting, which causes prediction accuracy to fall the most~\cite{ghorbani2019towards, vielhaben2023multi}. 
    \item STCE Concept Importance: it computes the importance rank of each concept for a given class~\cite{ji2023spatial}. 
    \item T-CAR score: it quantifies the proportion of instances belonging to a specific class whose representations fall within the positive concept region~\cite{crabbe2022concept}. 
    \item Sobol score: 
    it measures the contribution of a concept and its interactions of any order with any other concepts to the model output variance~\cite{fel2023craft}. 
    \item Concept Weight: in concept-based models employing a linear layer from concept to task, the same weight represents the importance of a concept for a given class~\cite{yuksekgonul2022posthoc, yang2023language}. 
    \item Concept Relevance: the predicted importance of a concept for a final class in a concept-based model, determined differently for each sample~\cite{alvarez2018towards,rigotti2022attentionbased, barbiero2023interpretable}. 
\end{itemize}

\myparagraphbold{(ii) Concept effect on task performance}
\begin{itemize}
    \item Completeness score: it measures the amount to which concept scores are sufficient for predicting model outcomes~\cite{yeh2020completeness}. This is based on the assumption that the concept scores of \textit{complete} concepts are sufficient statistics of the model prediction. 
    \item Fidelity: it counts the number of times in which the original and approximate model predictions are equal over the total number of images~\cite{zhang2021invertible, sarkar2022framework, fel2023craft}. 
    \item Faithfulness: this measures the predictive capacity of the generated concepts~\cite{sarkar2022framework}. It represents the capability of the overall concept vector to predict the ground truth task label. 
    \item Concept Efficiency: in concept-based models, the concept capacity to predict the task depends on the number of concepts employed~\cite{zarlenga2022concept}. Richer concept representations mitigate this problem. 
    \item Conciseness: similar to the concept efficiency but for post-hoc methods, it represents the number of concepts required to reach a certain grade of completeness and so a desired level of final accuracy~\cite{vielhaben2023multi}. 
\end{itemize}

\subsection{Quality of Concepts}

\myparagraphbold{i) Concepts properties}
 \begin{itemize}
    \item Mutual Information: it quantifies the amount of information retained in a concept representation with respect to the input data~\cite{zarlenga2022concept}. 
    \item Distinctiveness: it quantifies the distinction between different concepts by considering the extent of their coverage~\cite{wang2023learning}. 
    \item Completeness: Different from the previous definition of completeness given in~\cite{yeh2020completeness}, it measures how well a concept covers a specific associated class in the dataset~\cite{wang2023learning}. 
    \item Purity: it quantifies the ability to discover concepts covering only a single class~\cite{wang2023learning}. 
\end{itemize}

\myparagraphbold{(ii) Relation of Concepts with internal network representation}

\begin{itemize}
    \item Intersection over Union (IoU): it measures the degree of overlapping of the bounding boxes representing the concepts and the activation maps of the single node~\cite{bau2017network, fong2018net2vec, xuanyuan2023global} (and adopted~\cite{zhang2018interpretable} with the name of \textit{Part Interpretability}).  
    \item Location stability: it assesses the consistency in how a filter represents the same object part across various objects~\cite{zhang2018interpretable}. 
\end{itemize}

\myparagraphbold{(iii) Concept prediction error}

\begin{itemize}
    \item Concept Error: 
    it measures how close the concepts learned are to the concept ground truth~\cite{koh2020concept} (and used in~\cite{sarkar2022framework} with the name Explanation Error). It involves computing the L2 distance between the concepts learned and the ground truth concepts to measure the alignment. 
    \item AUC score: this score measures whether the samples belonging to a concept are ranked higher than others~\cite{chen2020concept}. That is, the AUC score indicates the purity of the concepts. 
    \item  Misprediction-overlap metric (MPO): it computes the sample fraction in the test set with at least \textit{m} relevant concepts predicted incorrectly~\cite{kazhdan2020now}. A larger MPO score implies a bigger proportion of incorrectly predicted concepts. 
\end{itemize}

\section{Methods Resources}
\label{app:resources}
In this section, we extend Section \ref{sec:evaluation}, employing an orthogonal direction of analysis. Rather than describing the singular resources, here we report tables \ref{tab:summary:posthoc2}, \ref{tab:summary:bydesign2} that contain, for each method, the type of resources employed in terms of metrics, dataset, and human evaluation, following the categorization provided in Section~\ref{sec:dimanalysis}.
\begin{table}[]
\caption{Post-hoc Concept-based Explainability methods. We characterize the approaches based on the following dimensions.
    The \textit{Code/Models availability, Data release}: (\vmark), no (\xmark).
    \textit{Human evaluation } (Human eval), if a user study is conducted for evaluation purposes: (\vmark), no (\xmark).
    Full category description in Section~\ref{sec:categorization}.}
    \label{tab:summary:posthoc2}
\begin{tabular}{l|lcccc}
 & \multicolumn{1}{c}{\textbf{Method}} & \textbf{\begin{tabular}[c]{@{}c@{}}Code/Mod. \\ Avail.\end{tabular}} & \textbf{\begin{tabular}[c]{@{}c@{}}Data \\ Release\end{tabular}} & \textbf{\begin{tabular}[c]{@{}c@{}}New \\ metric\end{tabular}} & \textbf{\begin{tabular}[c]{@{}c@{}}Human \\ eval\end{tabular}} \\ \hline
\multirow{9}{*}{\rotatebox{90}{Supervised}} 
& T-CAV   \cite{kim2018interpretability} & \vmark & \xmark & \vmark & \vmark \\
 & CAR \cite{crabbe2022concept} & \vmark & \xmark & \vmark & \xmark \\
 & IBD \cite{zhou2018interpretable} & \vmark & \xmark & \xmark & \vmark \\
 & CaCE \cite{goyal2019explaining} & \xmark & \xmark & \vmark & \xmark \\ 
& CPM \cite{wu2023causal}  & \vmark & \xmark & \xmark & \xmark \\ 
 & Obj. Det \cite{zhou2015object} & \xmark & \xmark &  \xmark & \vmark  \\ 
 & ND  \cite{bau2017network} & \vmark & \vmark & \vmark & \vmark \\
 & Net2Vec \cite{fong2018net2vec} & \vmark & \xmark & \xmark & \xmark \\
& GNN-CI \cite{xuanyuan2023global}  & \vmark & \xmark  & \xmark & \xmark \\ \hline
\multirow{7}{*}{\rotatebox{90}{Unsupervised}} 
& ACE \cite{ghorbani2019towards} & \vmark & \xmark & \xmark & \vmark \\
 & Compl. Aware \cite{yeh2020completeness}  & \vmark & \xmark  & \vmark  & \vmark  \\ 
 & ICE \cite{zhang2021invertible} & \vmark & \xmark & \xmark & \vmark \\
 & CRAFT   \cite{fel2023craft} & \vmark & \xmark & \vmark & \vmark \\ 
  & MCD \cite{vielhaben2023multi}  & \vmark & \xmark  & \xmark & \xmark  \\ 
   & DMA \& IMA \cite{leemann2023are}  & \vmark & \vmark  & \xmark & \xmark  \\ 
& STCE   \cite{ji2023spatial} & \vmark & \xmark & \xmark & \xmark \\ 

 \hline
\end{tabular}
\end{table}
\begin{table}[]
\caption{Explainable By-Design Concept-based Models. We characterize the approaches based on the following dimensions. 
     The \textit{Code/Models availability, Data release}: (\vmark), no (\xmark).
    \textit{Human evaluation } (Human eval), if a user study is conducted for evaluation purposes: (\vmark), no (\xmark).: (\vmark), no (\xmark).
    \textit{Human evaluation } (Human eval), if a user study is conducted for evaluation purposes: (\vmark), no (\xmark).
    Full category description in Section~\ref{sec:categorization}.}
    \label{tab:summary:bydesign2}
\begin{tabular}{l|l|cccc}
 & \textbf{Method} & \textbf{\begin{tabular}[c]{@{}c@{}}Code/Mod\\ Avail.\end{tabular}} & \textbf{\begin{tabular}[c]{@{}c@{}}Data \\ Release \end{tabular}} & \multicolumn{1}{l}{\textbf{\begin{tabular}[c]{@{}c@{}}New\\ Metric\end{tabular}}} & \multicolumn{1}{c}{\textbf{\begin{tabular}[c]{@{}c@{}}Human \\ eval\end{tabular}}} \\ \hline
\multirow{9}{*}{\rotatebox{90}{Supervised}} 
& CBM \cite{koh2020concept} & \vmark& \xmark& \xmark& \xmark\\
 & LEN \cite{barbiero2022entropy, ciravegna2023logic,   jain2022extending} & \vmark& \xmark& \xmark& \vmark\\
 & CEM\cite{zarlenga2022concept} & \vmark& \xmark& \vmark& \xmark\\
 & ProbCBM \cite{kim2023probabilistic} & \vmark& \xmark& \xmark& \xmark\\ 
 & DCR\cite{barbiero2023interpretable} & \vmark& \xmark& \xmark& \xmark\\
  & CW  \cite{chen2020concept} & \vmark& \xmark& \vmark& \xmark\\
  & CME \cite{kazhdan2020now} & \vmark& \vmark& \vmark& \xmark\\
   & PCBM  \cite{yuksekgonul2022posthoc} & \vmark& \xmark& \xmark& \xmark\\ 
 & CT  \cite{rigotti2022attentionbased} & \vmark& \xmark& \xmark& \xmark\\

\hline
\multirow{10}{*}{\rotatebox{90}{Unsupervised}} 
 & Interp. CNN \cite{zhang2018interpretable} & \vmark& \xmark& \xmark& \xmark\\ 
& SENN   \cite{alvarez2018towards} & \xmark& \xmark& \xmark& \xmark\\
 & SelfExplain   \cite{rajagopal2021selfexplain} & \vmark& \xmark& \xmark & \vmark\\
 & BotCL  \cite{wang2023learning} & \vmark& \xmark& \xmark& \vmark\\
 & PrototypeDL \cite{li2018deep} & \vmark& \xmark& \xmark& \xmark\\
 & ProtoPNet \cite{chen2019looks} & \vmark& \xmark& \xmark& \xmark\\ 
 & ProtoPool \cite{rymarczyk2022interpretable} & \vmark& \xmark& \xmark& \vmark\\ 
 & DeformableProtoPNet \cite{donnelly2022deformable} & \vmark& \xmark& \xmark& \xmark\\
 & HPnet \cite{hase2019interpretable} & \vmark& \xmark& \xmark& \xmark\\ 
 & ProtoPShare   \cite{rymarczyk2021protopshare} & \vmark & \xmark & \xmark & \vmark\\ \hline
\multirow{3}{*}{\begin{tabular}[c]{@{}l@{}}Hybrid \end{tabular}} 
& CBM-AUC \cite{sawada2022concept} & \xmark& \xmark& \xmark& \xmark\\
 & Ante-hoc expl. \cite{sarkar2022framework} & \vmark& \xmark& \xmark& \xmark\\ 
 & GlanceNets   \cite{marconato2022glancenets} & \vmark& \vmark & \xmark& \xmark\\ 
\hline
\multirow{2}{*}{Generative} 
& LaBO \cite{yang2023language} & \vmark& \xmark& \xmark& \vmark\\
 & Label-free CBM   \cite{oikarinen2023label} & \vmark& \xmark& \xmark& \xmark\\ \hline
\end{tabular}
\end{table}

\end{document}